\pdfoutput=1

\documentclass[11pt]{article}
\usepackage[dvipsnames,table]{xcolor}

\usepackage{ACL2023}

\usepackage{times}
\usepackage{latexsym}
\usepackage[T1]{fontenc}
\usepackage[utf8]{inputenc}
\usepackage{microtype}
\usepackage{inconsolata}
\usepackage{adjustbox}
\usepackage{tikz}
\usepackage{graphicx} 
\usepackage{multirow}
\usepackage{comment}
\usepackage{arydshln}
\usepackage{listings}
\usepackage{verbatimbox}

\usepackage[T1]{fontenc}

\usepackage{amssymb} 
\usepackage{pifont} 
\usepackage{arydshln} 

\usepackage{hyperref, relsize, import, amsmath, setspace, soul, booktabs, svg, array, xcolor, dblfloatfix, enumitem, cleveref, xfrac}

\newenvironment{hang}{
    \leftskip   8pt
    \rightskip  8pt
    \footnotesize
}{\par}

\newcommand{\inputoutput}[2]{\begin{hang}
\vspace{2pt}
\noindent \textbf{\textsc{#1}}:
\vspace{2pt}
\\
\noindent\theverbbox
\vspace{2pt}
\end{hang}}

\newenvironment{myquote}[1]%
  {\list{}{\leftmargin=#1\rightmargin=#1}\item[]}%
  {\endlist}

\definecolor{blue1}{HTML}{d1eeea}
\definecolor{blue2}{HTML}{a8dbd9}
\definecolor{blue3}{HTML}{85c4c9}
\definecolor{blue4}{HTML}{68abb8}
\definecolor{blue5}{HTML}{4f90a6}
\definecolor{blue6}{HTML}{3b738f}
\definecolor{blue7}{HTML}{2a5674}

\definecolor{red1}{HTML}{CB0E26}
\definecolor{red2}{HTML}{DA2323}
\definecolor{red3}{HTML}{E7464A}
\definecolor{red4}{HTML}{FF5429}
\definecolor{red5}{HTML}{fb6a4a}
\definecolor{red6}{HTML}{fc9272}
\definecolor{red7}{HTML}{fcbba1}
\definecolor{red8}{HTML}{fee0d2}
\definecolor{red9}{HTML}{fff5f0}

\definecolor{redf1}{HTML}{FCDEDE}

\definecolor{redhl}{HTML}{d7301f}

\definecolor{green1}{HTML}{1e7158}
\definecolor{green2}{HTML}{43b38e}
\definecolor{green3}{HTML}{36A353}

\newcommand{\greentriangle}{%
  \begin{tikzpicture}
    \filldraw[fill=white, draw=black, rotate=180] 
    (0,0) -- (0.1cm,0.2cm) -- (0.2cm,0) -- cycle;
  \end{tikzpicture}%
}

\definecolor{lightred}{RGB}{105,17,10}
\definecolor{lightblue}{RGB}{207, 215, 250}
\definecolor{darkred}{RGB}{105,17,10}
\definecolor{darkblue}{RGB}{0,0,139}

\usepackage[textsize=scriptsize]{todonotes}

\newcommand{\gradecolor}[1]{%
  \ifnum#1<30 \cellcolor{blue1}\color{black}%
  \else \ifnum#1<40 \cellcolor{blue2}\color{black}%
  \else \ifnum#1<50 \cellcolor{blue3}\color{black}%
  \else \ifnum#1<58 \cellcolor{blue4}\color{black}%
  \else \ifnum#1<66 \cellcolor{blue5}\color{white}%
  \else \ifnum#1<78 \cellcolor{blue6}\color{white}%
  \else \cellcolor{blue7}\color{white}%
  \fi\fi\fi\fi\fi\fi%
}

\newcommand{\newcellcolor}[2]{%
  \cellcolor{#1}\color{#2}%
}
\newcommand{\aspace}{\hspace{0.4em}}
\newcommand{\gt}{$^{\pi}$}
\newcommand{\cmu}{$^{e}$}

\title{Reducing Privacy Risks in Online Self-Disclosures \\with Language Models}

\author{
  Yao Dou\gt \aspace
  Isadora Krsek\cmu \aspace
  Tarek Naous\gt \aspace 
  Anubha Kabra\cmu \\
  \textbf{Sauvik Das}\cmu \aspace 
  \textbf{Alan Ritter}\gt \aspace
  \textbf{Wei Xu}\gt \\
  \gt Georgia Institute of Technology \aspace 
  \cmu Carnegie Mellon University \\
  \texttt{douy@gatech.edu}
  }

\begin{document}
\maketitle
\begin{abstract}
Self-disclosure, while being common and rewarding in social media interaction, also poses privacy risks. In this paper, we take the initiative to protect the user-side privacy associated with online self-disclosure through \textit{detection} and \textit{abstraction}. We develop a taxonomy of 19 self-disclosure categories and curate a large corpus consisting of 4.8K annotated disclosure spans. We then fine-tune a language model for detection, achieving over 65\% partial span F$_1$. We further conduct an HCI user study, with 82\% of participants viewing the model positively, highlighting its real-world applicability. Motivated by the user feedback, we introduce the task of self-disclosure abstraction, which is rephrasing disclosures into less specific terms while preserving their utility, e.g., \textit{``Im 16F''} to \textit{``I'm a teenage girl''}. We explore various fine-tuning strategies, and our best model can generate diverse abstractions that moderately reduce privacy risks while maintaining high utility according to human evaluation. To help users in deciding which disclosures to abstract, we present a task of rating their importance for context understanding. Our fine-tuned model achieves 80\% accuracy, on par with GPT-3.5. Given safety and privacy considerations, we will only release our corpus and models to researchers who agree to the ethical guidelines outlined in our Ethics Statement.\footnote{To request access, please send us an email and submit a request for the
\begingroup
\hypersetup{urlcolor=red}
\href{https://huggingface.co/datasets/douy/reddit-self-disclosure}{corpus}, \href{https://huggingface.co/douy/deberta-v3-large-self-disclosure-detection}{detection model}, \href{https://huggingface.co/douy/Llama-2-7B-lora-instruction-ft-abstraction-three-span}{abstraction model}.
\endgroup}

\end{abstract}

\section{Introduction}
Self-disclosure --- \textit{the communication of personal information to others} \cite{jourard1971self,Cozby1973SelfdisclosureAL} --- is prevalent in online public discourse.
Disclosing personal information allows users to seek social support, build community, solicit context-specific advice, and explore aspects of their identity that they feel unsafe exploring offline \cite{luo2020self}.
Consider the following (hypothetical, but representative) Reddit post:
\begin{myquote}{0.25in}
 \vspace{-0.3em}
\textit{Im 16F I think I want to be a bi M}
 \vspace{-0.3em}
\end{myquote}

\begin{figure}[!t]
  \centering
    \includegraphics[width=0.96\linewidth]{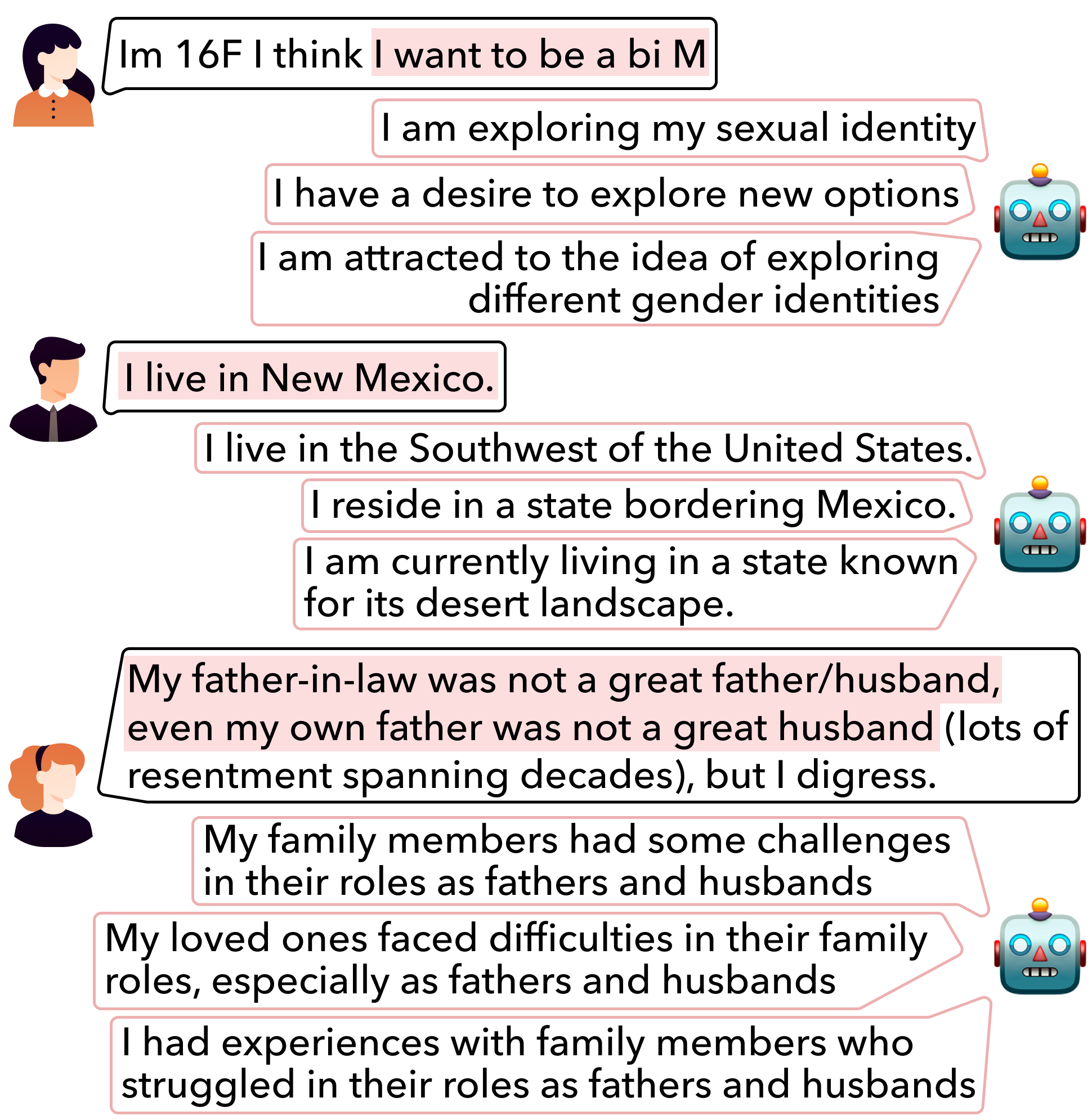}
    \vspace{-5pt}
 \caption{Our model can provide diverse abstractions for \sethlcolor{redf1}\hl{self-disclosures} of any length to suit user preferences. This approach effectively reduces privacy risks without losing the essence of the message.}
    \label{fig:figure1}
    \vspace{-12pt}
\end{figure}

\noindent The author discloses their age, gender, and sexual orientation to express themselves. However, these self-disclosures simultaneously expose them to privacy risks, notably regret of the disclosure \cite{sleeper2016everyday} and doxxing \cite{staab2023beyond}, which are particularly acute for marginalized populations \cite{lerner2020privacy}. This raises a critical question: \textbf{How can we help users identify and mitigate privacy risks in online self-disclosures?}

Prior works on self-disclosure \cite[\textit{inter alia}]{valizadeh2021identifying,cho2022assessing,staab2023beyond} and anonymization tools \cite{lison-etal-2021-anonymisation} focus on only a limited set of self-disclosures (e.g., health
issues) or inferring personal attributes (aka. user profiling), often at sentence/post levels. They do not pinpoint the exact words of disclosures in the sentence, nor have broad enough coverage of different kinds of disclosures. Both are crucial for real-world users to take control of what they want to disclose and protect their privacy.

\begin{table*}[t!]
\setlength{\tabcolsep}{2pt}
\resizebox{\linewidth}{!}{%
\renewcommand{\arraystretch}{1.04}
\begin{tabular}{lccl}
\toprule
\textbf{Category} & \textbf{\#Spans} & \multicolumn{1}{c}{\textbf{Avg Len}} & \textbf{Example} \\ \midrule
\textit{\textbf{Demographic Attributes}} \\
\hspace{0.9em}{\sc Location} & 525 & 5.70{\small$\pm3.85$} & \sethlcolor{redf1}\hl{I live in the UK} and a diagnosis is really expensive, even with health insurance \\
\hspace{0.9em}{\sc Age} & 308 & 2.93{\small$\pm1.72$} & \sethlcolor{redf1}\hl{I am a 23-year-old} who is currently going through the last leg of undergraduate school \\
\hspace{0.9em}{\sc Relationship Status} & 287 & 6.72{\small$\pm5.97$} & \sethlcolor{redf1}\hl{My partner} has not helped at all, and I’m bed ridden now \\
\hspace{0.9em}{\sc Age/Gender} & 248 & 1.42{\small$\pm0.71$} & For some context, I \sethlcolor{redf1}\hl{(20F)}, still live with my parents \\
\hspace{0.9em}{\sc Pet} & 192 & 6.93{\small$\pm7.31$} & Hi, \sethlcolor{redf1}\hl{I have two musk turtles} and have never had any health problems before at all \\
\hspace{0.9em}{\sc Appearance} & 173 & 6.96{\small$\pm6.25$} & Same here. \sethlcolor{redf1}\hl{I am 6'2}. No one can sit behind me. \\
\hspace{0.9em}{\sc Husband/BF} & 148 & 6.89 {\small$\pm7.24$} & \sethlcolor{redf1}\hl{My husband} and I vote for different parties \\
\hspace{0.9em}{\sc Wife/GF} & 144 & 5.24{\small$\pm4.42$} & \sethlcolor{redf1}\hl{My gf} and I applied, we're new but fairly active! \\
\hspace{0.9em}{\sc Gender} & 110 & 3.28{\small$\pm3.10$} & Am I insane? Eh. \sethlcolor{redf1}\hl{I'm just a girl} who wants to look on the outside how I feel on the inside. \\
\hspace{0.9em}{\sc Race/Nationality} & 99 & 3.63{\small$\pm2.37$} & \sethlcolor{redf1}\hl{As Italian} I hope tonight you will won the world cup \\
\hspace{0.9em}{\sc Sexual Orientation} & 58 & 6.52{\small$\pm7.47$} & \sethlcolor{redf1}\hl{I'm a straight man} but I do wanna say this \\
\hspace{0.9em}{\sc Name} & 21 & 3.81{\small$\pm3.48$} & Hello guys, \sethlcolor{redf1}\hl{my name is xxx} and I love  travelling \\
\hspace{0.9em}{\sc Contact} & 14 & 5.69{\small$\pm3.56$} & \sethlcolor{redf1}\hl{xxx is my ig} \\
\hdashline[2pt/2pt]
\addlinespace[2.4pt]
\textit{\textbf{Personal Experiences}} \\
\hspace{0.9em}{\sc Health} & 783 & 10.36{\small$\pm9.78$} & \sethlcolor{redf1}\hl{I am pretty sure I have autism}, but I don’t want to get an official diagnosis. \\
\hspace{0.9em}{\sc Family} & 543 & 9.27{\small$\pm8.73$} & \sethlcolor{redf1}\hl{My little brother (9M)} is my pride and joy \\
\hspace{0.9em}{\sc Occupation} & 428 & 8.90{\small$\pm6.60$} & \sethlcolor{redf1}\hl{I'm a motorcycle tourer (by profession)}, but when I'm off the saddle I'm mostly bored \\
\hspace{0.9em}{\sc Mental Health} & 285 & 16.86{\small$\pm16.28$} & I get asked this pretty regularly.. but \sethlcolor{redf1}\hl{I struggle with depression and ADHD} \\
\hspace{0.9em}{\sc Education} & 229 & 9.92{\small$\pm7.71$} & Hi there, \sethlcolor{redf1}\hl{I got accepted to UCLA (IS)}, which I'm pumped about. \\
\hspace{0.9em}{\sc Finance} & 153 & 12.00{\small$\pm9.19$} & Yes. \sethlcolor{redf1}\hl{I was making \$68k a year and had around \$19k in debt} \\ \bottomrule
\end{tabular}
}
\caption{Statistics and examples for each self-disclosure category in our dataset, sorted by decreasing frequency. Personal identifiable information are redacted as `xxx' to be shown here.}
\label{tab:disclosure-categories}
\vspace{-10pt}
\end{table*}

In this work, we take the important first steps in protecting user-side privacy with broad-coverage \textbf{self-disclosure detection} and \textbf{abstraction}. Our models are extensive in capturing 19 categories of disclosure (Table \ref{tab:disclosure-categories}). We use a human-centered, iterative design process with actual end-users to evaluate and improve the models. Our detection model helps users scrutinize their contents to the word level (e.g., \textit{``16F''}) to account for privacy risks, while our abstraction model assists them in rephrasing their content to reduce these risks.

Specifically, we introduce a comprehensive taxonomy for self-disclosure (Table \ref{tab:disclosure-categories}) that consists of 13 demographic attributes and 6 personal experiences. We create a high-quality dataset with human annotations on 2.4K Reddit posts, covering 4.8K varied self-disclosures. With this corpus, we fine-tune a language model to identify the self-disclosures in the given text, achieving over 65\% partial span F$_1$. Besides the standard NLP evaluations, more importantly, we conducted an HCI user study with 21 Reddit users to validate the real-world applicability. 82\% participants have a positive outlook on the model, while also providing valuable suggestions on aspects such as personalization and explainability, which are often overlooked in benchmark assessments. 

Participants also express a need for a tool that can (quote) \textit{``rewrite disclosures for me in a way that I don't worry about privacy concerns''}. We thus introduce the novel task of self-disclosure abstraction, with the goal of rephrasing disclosure spans into less specific words without losing the utility or essence of the message.
For example, providing three alternatives such as \textit{``I am exploring my sexual identity''} in place of  \textit{``I want to be a bi(sexual) M(an)''} to let users choose based on their preferences, see Figure \ref{fig:figure1}.
We showcase the effectiveness and uniqueness of our task in comparison to other related tasks such as paraphrasing and sentence-level abstraction.
We also experiment with different fine-tuning strategies. The best model, distilled on GPT-3.5 generated abstractions, can increase privacy moderately (scoring 3.2 out of 5 with 5 being the highest level of detail removal) while preserving high utility (scoring 4 out of 5). The model's abstractions are also very diverse, offering varied expressions (scoring 4.6 out of 5).\footnote{These numbers are Likert-scale of the human eval.}
To assist users in determining which disclosures to abstract, we additionally present the task of rating the importance of self-disclosure in understanding the context. Our fine-tuned model achieves comparable performance to GPT-3.5 of 80\% accuracy.

In summary, our key contributions include:
\begin{itemize}
[itemsep=-3pt,topsep=1pt,leftmargin=8pt]
    \item We introduce a new corpus annotated for self-disclosure with 19 categories (\S \ref{sec:corpus}).
    \item We conduct a study with real Reddit users and show that our detection model helps users manage privacy risks (\S \ref{sec:user_study}).
    \item Motivated by the user study, we propose a novel self-disclosure abstraction task, and show promising model results in human evaluations (\S \ref{sec:abstraction}).
\end{itemize}

\section{Fine-grained Self-Disclosure Detection}
\label{sec:corpus}
To mitigate privacy leaks and alert people about their self-disclosures, it is essential to highlight specific text segments that disclose personal information, rather than simply classifying sentences as containing disclosures. To cover a wide spectrum of disclosures, we design a detailed taxonomy of 19 self-disclosure categories, which is more extensive than prior work (see Related Work in \S \ref{sec:relatedwork}). We further construct an annotated corpus for training automatic models to detect self-disclosures at word-level, which supports our user study in \S \ref{sec:user_study}.

\subsection{Annotated Self-Disclosure Corpus}
\label{subsec:corpus}

We curate a large dataset of 2.4K Reddit posts manually annotated with 4.8K self-disclosure spans.

\paragraph{Data Collection.} We use the public Reddit data dump from December 2022, which contained 35.86M posts. We filter out 42.52\% of posts that were marked as ``NSFW'' or ``Over\_18'', indicating adult content, as well as those that were removed by moderators. We keep only English posts as determined with a probability above 0.7 by the fastText \cite{joulin2016fasttext} language identifier.\footnote{\url{https://fasttext.cc/}} This results in a total of 4.01M posts, from which we randomly sample 10K and then reconstruct the full post threads with all comments and reply chains for each post via the Reddit API. We then ask two annotators to review the 10K posts on whether containing self-disclosures, culminating in a set of 2,415 posts for subsequent span annotation.

\paragraph{Self-Disclosure Taxonomy.} Different from prior work that focuses on specific types of self-disclosure (e.g., health \cite{valizadeh-etal-2021-identifying} or sexual harassment \cite{chowdhury2019speak}), we categorizing disclosures into 19 types that are commonly shared by social media users. The taxonomy is refined iteratively through three rounds of pilot studies. Table \ref{tab:disclosure-categories} presents statistics and examples for these 19 categories that fall into two main groups: \textit{demographic attributes} and \textit{personal experiences}. Attributes refer to static personal characteristics that are often stated succinctly such as \texttt{name}, \texttt{age}, and \texttt{gender}. Experiences, on the other hand, relate to events that an individual engages in over time, which are more complex and dynamic, such as \texttt{health} and \texttt{education}. For disclosures concerning others, such as family members, we direct annotators to label them under a general category (i.e., \texttt{family}).

\paragraph{Annotation Process.} To ensure quality and privacy standards, we hire seven in-house annotators who were given training tutorials and 20 annotation exercise examples. We ask annotators to highlight text spans that reveal personal information within each post (including comments) and categorize them into one of 19 self-disclosure types. To enhance accuracy and relevance of self-disclosures detection, we instruct annotators to select spans with contextual information, which provides more nuanced training signals for models. For example, \textit{``\sethlcolor{redf1}\hl{I live in the US}}'' would be preferred over a minimal span like \textit{``\sethlcolor{redf1}\hl{US}''}, which is isolated from its self-referential context. The annotation process was organized into 10 batches, with the final two batches--comprising the most recent posts--undergoing a double annotation process followed by adjudication, which we will use for continual fine-tuning and evaluation. The inter-annotator agreement is 0.54 by Krippendorff's $\alpha$ \cite{krippendorff2018content}; see agreement by category in Appendix \ref{app:iaa}.

\begin{table}[t!]
  \centering
  \setlength{\tabcolsep}{3pt}
  \resizebox{0.99\columnwidth}{!}{
    \begin{tabular}{lccc}
      \toprule
      \textbf{Class (\#spans)} & \textbf{RoBERTa} & \textbf{DeBERTa} & \textbf{GPT-4} \\
      \midrule
      \textsc{Age (35)}  & 72.46 & 70.77 & \textbf{80.0} \\
      \textsc{Age\&Gender (17)}  & \textbf{84.21} & 70.27 & 74.42  \\
      \textsc{Race/Nationality (8)} & 60.0 & \textbf{82.35} & 70.59 \\
      \textsc{Gender (17)} & 61.11 & \textbf{72.73} & 57.14  \\
      \textsc{Location (41)} & 71.26 & \textbf{73.33} & 54.35 \\
      \textsc{Appearance (31)} & 64.41 & \textbf{67.74} & 42.55 \\
      \textsc{Wife/GF (30)} & 66.67 & \textbf{75.86} & 64.52 \\
      \textsc{Finance (33)} & 68.66 & \textbf{71.43} & 54.55 \\
       \textsc{Occupation (44)} & 64.44 & \textbf{65.22} & 52.75  \\
       \textsc{Family (44)}  & \textbf{58.70} & 49.02 & 58.25  \\
      \textsc{Health (40)} &  56.84 & \textbf{58.82} & 38.02 \\
      \textsc{Mental Health (46)} & \textbf{64.71} & 63.16 & 52.73 \\
      \textsc{Husband/BF (14)} & \textbf{75.0} & 70.59 & 68.97 \\
        \textsc{Education (21)}  & 68.09 & \textbf{69.23} & 51.06  \\
      \textsc{Pet (15)} & 46.15 & \textbf{55.17} & 48.28  \\
      \textsc{Relation. Status (31)} & 41.10 & \textbf{43.08} & 42.86  \\
      \textsc{Sexual Orient. (12)} & \textbf{76.19} & 58.33 & 69.57 \\
      \midrule
      \textsc{Average} & 64.71 & \textbf{65.71} & 57.68 \\
      \bottomrule
    \end{tabular}}
  \caption{Test performance per class in partial F1  for fine-tuned models and prompted \texttt{GPT-4-0125-preview}.}
  \vspace{-10pt}
  \label{table:performance_by_model}
\end{table}

\subsection{Automatic Self-Disclosure Detection}
\label{sec:id_model}

With our annotated corpus, we fine-tune RoBERTa-large \cite{liu2019roberta} and DeBERTaV3-large \cite{he2021debertav3} to detect self-disclosures as a sequence tagging task. The models are first fine-tuned on 4,959 sentences with single annotations, and continually trained on 802 sentences with adjudication annotations. In total, there are 5,761/218/400 sentences for train/val/test.
We also evaluate prompted GPT-4 Turbo for comparison, but it is important to note that prompting is less practical  due to higher costs and inefficiencies, and the privacy-sensitive nature of this task might lead users to prefer models that can operate on their local devices.
We report partial span-level F$_1$, a middle ground that is stricter than token-level but more lenient on span boundaries than full span-level F$_1$.
It considers a predicted span as correct if it contains or is contained by a reference span, with the overlap more than 50\% of the longer span's length.
Table \ref{table:performance_by_model} presents the test set performance. Fine-tuned DeBERTa performs the best with large margin ahead GPT-4 Turbo, aligning with previous findings on prompted LLMs' low performances on span-level tasks \cite{ashok2023promptner,staab2023beyond}.
For infrequent categories: \texttt{name} and \texttt{contact} categories, we combine a sentence classifier trained on our data (determine whether a sentence contains a self-disclosure) with an existing NER model \cite{yamada-etal-2020-luke} and regular expressions.
Additional details, including token and span-level F1, and binary classification results are provided in Appendix \ref{app:further_results}.

\vspace{2pt}
\noindent \textbf{Identifying Self-Disclosures in ShareGPT.}
As LLM-based chatbots, such as ChatGPT, demonstrate more advanced capabilities, many internet users make use of them to assist with daily tasks. Users may share personal information during these interactions, such as seeking help to revise resumes. Since these conversations could be stored by service providers for future training, this poses a risk of privacy leakage through data memorization \cite{carlini2022quantifying}. So we test whether our DeBERTa model is able to detect self-disclosures in conversations with ChatGPT.
We randomly sampled 1,600 human-authored conversation turns from ShareGPT,\footnote{\url{https://huggingface.co/datasets/anon8231489123/ShareGPT_Vicuna_unfiltered}} and, after a filtration and annotation process, obtained 105 turns with human-annotated self-disclosures.
We find \texttt{occupation} and \texttt{location} are most common, occurring 75 and 31 times, followed by \texttt{education}, \texttt{relationship} and \texttt{family} with around 8 times. Other categories occur less than 3 times. The average partial F$_1$ for these 5 categories is 60.64, slightly higher than in-domain performance of 59.98, demonstrating our model's generalizability. See Appendix \ref{app:sharegpt} for more details.

\section{User Study}
\label{sec:user_study}
To understand how real users think about our disclosure detection model for protecting their privacy, we recruited 21 Reddit users through Prolific for an interview study--a step that differentiates our approach from prior disclosure identification work. This user study and its analysis were led by three authors with expertise in HCI, privacy, and NLP.

\subsection{Participants and Study Design}

All participants recruited were aged 18 or older, had an active Reddit account, and had made at least three posts. After completing a screening survey, eligible participants were asked to fill out a pre-study survey, including a digital copy of the consent form describing the nature of the interview. After consenting, they were prompted to schedule an interview with researchers. Interviews took place over Zoom, averaging about 2 hours, during which participants were asked to share one of their Reddit posts that raised privacy concerns, and also to write a post that they were hesitant to publish for similar reasons.
More details on participants recruitment are provided in Appendix \ref{app:recruitment}. 
We then ran those posts through both our binary and multi-label models (\S \ref{sec:id_model}) and provided the annotated images of users' posts that display the detected self-disclosure spans to users. We asked participants about where they agreed and disagreed with model outputs, their overall impression of the model, if and how they would like to use the model outside of the study, as well as suggestions for improvement. Our study design was approved by the university's institutional review board (IRB).

\subsection{User Perceptions of Our Model}

In all, we see a significant majority (82\%) of participants had a positive outlook on the model. In addition, the multi-class model that highlighted disclosure categories was helpful to around  48\% of participants, aiding them in recognizing and understanding potential privacy risks in their posts.

More specifically, 62\% of participants expressed a desire to use it on their own posts, and another 10\% felt that even though they would see no need for such a tool themselves, they would recommend others they know to use this tool and suggest that it might be \textit{``a good idea for... kids and teens, like people who are new to the Internet.''} One participant said that \textit{``It would be interesting to run it through before I post something that I'm like nervous about and just see what it thinks and see if there are any areas where I can fix to make it less specific to me.''} An additional 10\% of participants mentioned that they would use it if they were more prone to making self-disclosures or if the model is further improved (more in \S \ref{subsec:id_limitations} and \S \ref{sec:abstraction}).

\subsection{User Feedback and New Opportunities}
\label{subsec:id_limitations}

When discussing why they viewed the model favorably, 62\% of participants mentioned the focus on word-level disclosures, 57\% mentioned self-reflection, and 48\% mentioned fine-grained categories. Users also provided feedback for improvement centering around accuracy, personalization, and desire for support in mitigating privacy risks.

One interesting finding from the user study is the divergence between annotators and real users in terms of what they think should be highlighted. Our initial design goal was to mark anything that the model identified as a self-disclosure to allow users to make informed decisions themselves. This approach led 4 out of the 21 participants to the believe that the model was ``over-sensitive'' and inaccurate because it highlighted content that participants did not believe was risky. This issue was succinctly summarized by one participant: \textit{``sometimes it's so oversensitive that it'll highlight things again (and again), and people might not use it because they get kind of fed up and irritated.''} We propose to address this problem by importance rating in \S \ref{sec:importance}.

Another interesting finding is that some participants suggested having the model account for their use of privacy-preserving strategies, e.g., when users intentionally author posts with false personal information, highlighting such information as a disclosure risk is not useful. Future model iterations could include features that allow users to adapt outputs to align with these strategies. One example could be proactively offering suggestions for altering text, potentially through strategic falsehoods that retain semantic utility. In fact, 24\% of participants sought recommendations on how to rephrase text spans that the model detected as a sensitive disclosure.
One participant articulated this need by stating: \textit{``could you rewrite this for me in a way that I don't have to worry about privacy concerns?''}
This feedback led us to explore methods to generate alternative phrasings of privacy-sensitive text spans, as we discuss next in \S \ref{sec:abstraction}.

\section{Self-Disclosure Abstraction}
\label{sec:abstraction}
Building upon insights from our user study, we introduce a novel task, self-disclosure abstraction, which rephrases disclosures with less specific details while preserving the content utility (see examples in Figure \ref{fig:figure1}).

\subsection{Task Definition}
Given a disclosure span within a sentence, the objective is to reduce sensitive and specific details while preserving the core meaning and utility. For example, in the sentence: \textit{`` \sethlcolor{redf1}\hl{I just turned 32 last month}} and have been really ...'', the highlighted disclosure span can be abstracted to \textit{``\sethlcolor{redf1}\hl{I recently entered my early 30s}''}. This task operates at the span level, functioning similarly to a text editing tool such as Grammarly. In practice, we envision that abstraction will work with the detection model in a pipeline--first identifying self-disclosures, then users can select which disclosures to abstract. Abstracted spans must fit seamlessly into the original sentence without changing the rest of the text.

\begin{table}[t!]
  \centering
  \setlength{\tabcolsep}{3pt}
  \resizebox{0.99\columnwidth}{!}{
    \begin{tabular}{lccc}
        \toprule
        \multirow{2}{*}{\textbf{Task}} & \textbf{Maintain} &  \textbf{Improve} & \textbf{Keep} \\
        & \textbf{Utility} & \textbf{Privacy} & \textbf{Surrounding Text} \\
        \midrule
        Sentence Paraphrasing & \ding{51} & \ding{55} & \ding{51} \\ 
        Sentence Abstraction & \ding{51} & \ding{51} & \ding{55} \\
        Span Abstraction & \ding{51} & \ding{51} & \ding{51} \\
        \midrule  \midrule
        \textit{Example} \\
        \multicolumn{4}{l}{\textbf{Sentence:} ``\underline{Not 21} so can't even drink really even tho \underline{I'm in Korea.}''} \\
        \multicolumn{4}{l}{\textbf{Sentence Paraphrasing:} ``Even though I'm in Korea, I can't actually} \\
        \multicolumn{4}{l}{drink because I'm not 21 yet.''} \\
        \multicolumn{4}{l}{\textbf{Sentence Abstraction:} ``Not old enough to legally consume alcohol} \\
        \multicolumn{4}{l}{even though I'm abroad.''} \\
        \multicolumn{4}{l}{\textbf{Span Abstraction:} ``Not of legal drinking age so can't even drink } \\
        \multicolumn{4}{l}{really even tho I'm abroad.''} \\
        \bottomrule
    \end{tabular}}
  \caption{Task comparison with an illustrative example.}
  \vspace{-10pt}
  \label{table:abstraction_tasks_comparison}
\end{table}

\begin{table*}[t!]
  \centering
  \setlength{\tabcolsep}{4pt}
  \renewcommand{\arraystretch}{1.15}
  \resizebox{0.99\textwidth}{!}{
    \begin{tabular}{lcccccccr}
      \toprule
    & \multicolumn{4}{c}{\textbf{Automatic Evaluation (Matching Metric)}} & \multicolumn{4}{c}{\textbf{Human Evaluation}} \\ \cmidrule(lr{1.2em}){2-5}  \cmidrule(lr{0.6em}){6-9} 
    \textbf{Methods} & BLEU & ROUGE-2 & ROUGE-L & Diversity & Rank Dist. (1$\rightarrow$9) & Rank \greentriangle & Raw & Z-Score \\
    \midrule
    \textit{Sampling} \\
      \hspace{5pt} Special Token & 17.40 & 18.81 & 38.42 & 2,576 &  \includegraphics[width=0.2\linewidth]{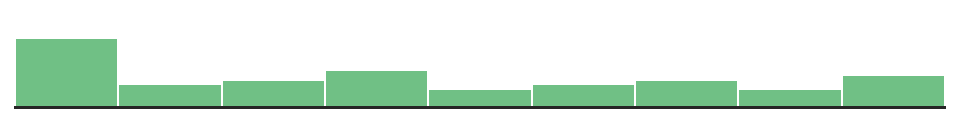} & 5.37 & 79.22 &  -0.10 \\
      \hspace{5pt} Instruction & 16.95 & 18.78 & 38.35 & 2,564 & \includegraphics[width=0.2\linewidth]{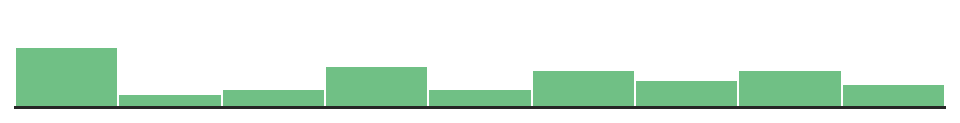} & 5.80 & 77.17 & -0.22 \\
      \hspace{5pt} Instruction (w/ thought) & 17.06 & 18.47 & 38.43 & 2,882 & \includegraphics[width=0.2\linewidth]{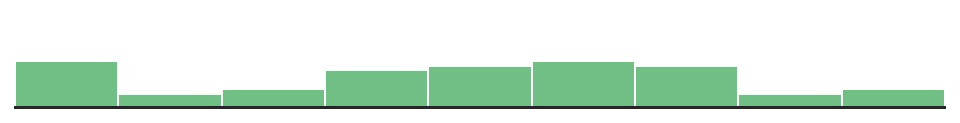} & 5.32 & 78.13 & -0.13 \\ \addlinespace[4pt] 
      \textit{End-to-end training} \\
      \hspace{5pt} Special Token & 17.22 & 19.67 & 38.38 & 2,911  & \includegraphics[width=0.2\linewidth]{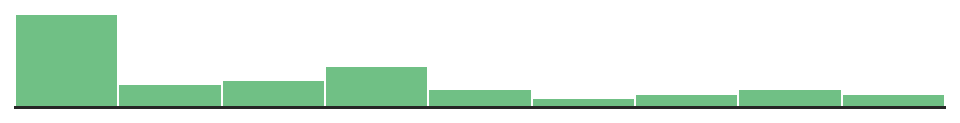} & 4.65 & 79.03 & -0.07 \\
      \hspace{5pt} Instruction & \underline{17.99} & 19.60 & \textbf{39.57} & \underline{2,992} & \includegraphics[width=0.2\linewidth]{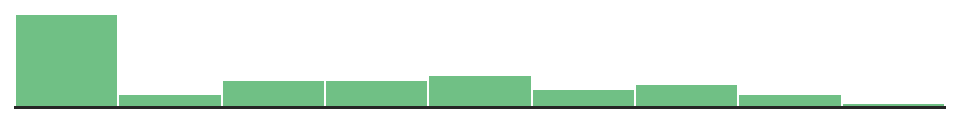} & \textbf{3.77} & \textbf{81.53} & \textbf{0.21} \\
      \hspace{5pt} Instruction (w/ thought) & 16.53 & 19.24 & 38.81 & 2,801 & \includegraphics[width=0.2\linewidth]{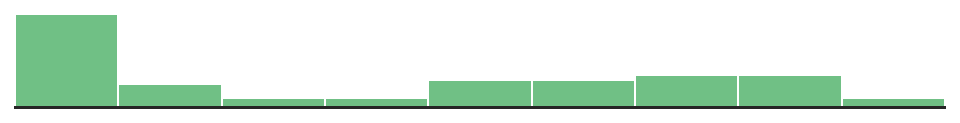} & 4.48 & 79.76 & 0.04 \\ \addlinespace[4pt]
     \textit{Iterative generation} \\
      \hspace{5pt} Special Token & \textbf{18.13} & \underline{19.74} & 38.71 & 2,913 & \includegraphics[width=0.2\linewidth]{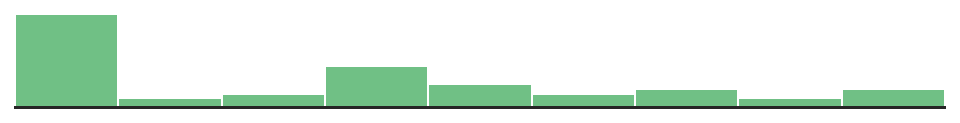} & \underline{4.12} & 80.72 & 0.11 \\
      \hspace{5pt} Instruction & 17.80 & \textbf{19.81} & \underline{39.56} & \textbf{3,067} & \includegraphics[width=0.2\linewidth]{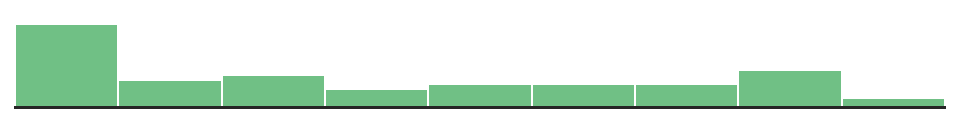} & 5.10 & 78.46 & -0.03 \\
      \hspace{5pt} Instruction (w/ thought) & 16.89 & 18.84 & 38.31 & 2,914 & \includegraphics[width=0.2\linewidth]{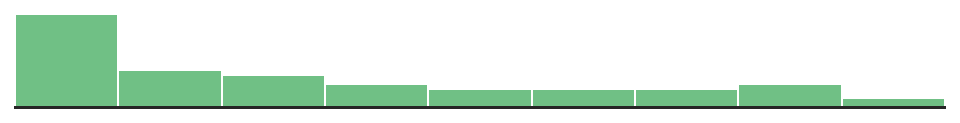} & 4.18 & \underline{81.27} & \underline{0.17} \\
      \bottomrule
    \end{tabular}}
  \caption{Test results on generating three alternative abstracted spans. End-to-end instruction tuning and iterative instruction training with thought achieve the top two performances under human evaluation. \textit{Rank Dist.} presents the histograms of the rank distribution, where 1 is the best and 9 the worst. \textit{Diversity} presents \# unique bigrams.}
  \vspace{-10pt}
  \label{table:generalization_three_spans_performance}
\end{table*}

\vspace{2pt}
\noindent  \textbf{Comparison with Sentence-level Tasks.} Table \ref{table:abstraction_tasks_comparison} illustrates the differences between sentential paraphrasing, sentence- and span-level abstraction. Sentential abstraction generalizes the entire sentence with or without disclosure spans provided. This approach generally modifies non-disclosure text as well as disclosures, affecting the original writing style, and potentially introducing unintended abstractions or hallucinations \cite{zhang2023siren}, which may be undesirable.
To assess the effectiveness of each task, we randomly sample 100 test sentences and apply each method by zero-shot prompting GPT-4. Two annotators are asked to rank and rate model outputs on a scale from 0 to 100, with an average 0.52 Kendall's $\tau$ for agreement. We aggregate the annotations by re-ranking the sums. Table \ref{table:abstraction_tasks_ranking} shows that when disclosures spans are provided, sentence-level abstraction achieves similar high effectiveness as span-level abstraction, with scores over 85.
Prompts are listed in Appendix \ref{app:prompt-template}.

\begin{table}[t!]
  \centering
  \setlength{\tabcolsep}{3pt}
  \resizebox{0.99\columnwidth}{!}{
    \begin{tabular}{lccccc}
        \toprule
        \textbf{Task} & \textbf{Rating} & \textbf{\textcolor{blue7}{Rank 1}} & \textbf{\textcolor{blue4}{Rank 2}} & \textbf{\textcolor{red6}{Rank 3}} & \textbf{\textcolor{red3}{Rank 4}} \\
        \midrule
        Sentence Paraphrasing & 72.52 & 4\% & 14\% & 16\% & 66\% \\
        \addlinespace[0.7mm]
        \multicolumn{4}{l}{Sentence Abstraction:} \\
        \hspace{4mm} w/o providing spans & 80.33 & 38\% & \textbf{34\%} & 22\% & 6\% \\
        \hspace{4mm} w/ providing spans & \textbf{86.16} & \textbf{54\%} & 24\% & 18\% & \textbf{4\%} \\
        \addlinespace[0.7mm]
        Span Abstraction & 85.62 & 50\% & 30\% & \textbf{14\%} & 6\% \\
        \bottomrule
    \end{tabular}}
  \caption{Average human rating and rank distribution across four tasks, evaluating overall effectiveness, which considers both utility preservation and privacy increase.}
  \vspace{-10pt}
  \label{table:abstraction_tasks_ranking}
\end{table}

\subsection{Automatic Generation of Training Data}
We use the most recent 10\% of posts (plus associated comments) from our corpus (\S\ref{subsec:corpus}), which are further divided into training (159 posts), dev (25 posts), and test sets (50 posts). While creating diverse abstractions is challenging for human annotators, LLMs are adept at this task. We use chain-of-thought \cite{wei2022chain} prompting with few-shot demonstrations, which asks the model to first generate a rationale on why the disclosure span needs abstraction, and then generate three diverse abstractions, aiming to accommodate varied user preferences. For training and dev sets, we choose GPT-3.5 (06-13) to balance cost and performance. We use GPT-4 (06-13) for the test set, given its more advanced capabilities. We list the prompt in Appendix \ref{app:prompt-template}.

\subsection{Abstraction Models}
\label{subsec:abstraction_models}

We fine-tune Llama-2-7B \cite{touvron2023llama} with LoRA \cite{hu2021lora} for generating abstractions.
We experiment with three different methods: sampling three times from an abstraction model (see Appendix \ref{app:further_abstraction_results}) that generates only one abstraction at a time (\textbf{sampling}), training an abstraction model that generates three abstractions all at once (input $\rightarrow$ A, B, C)  (\textbf{end-to-end training}), or using an iterative approach that breaks the three abstractions (A, B, C) into three separate training instances: input $\rightarrow$ A, input+A $\rightarrow$ B, input+A+B $\rightarrow$ C (\textbf{iterative generation}).
For each method, we consider the top three input-output setups identified in Appendix \ref{app:further_abstraction_results}, where we train models to generate only a single abstraction. These setups include formatting the input with special tokens and calculating loss on abstraction, formatting input with instruction and calculating loss either on abstraction or on rationale plus abstraction.
Overall, we evaluate a total of 9 models.

\subsection{Results}  

\noindent \textbf{Automatic Evaluation.} Table \ref{table:generalization_three_spans_performance} presents the test set performance for each model. We adopt matching BLEU and ROUGE metrics proposed by \citep{dou2021multitalk} to encourage diversity in generated abstractions. These matching metrics use the Hungarian algorithm \cite{kuhn1955hungarian} to calculate the highest matching scores among one-to-one pairings between generations and references. According to the automatic metrics, all three methods show similarly high performance, while generating three abstractions in iterative steps yields slightly better results.

\begin{figure}[!t]
  \centering
    \includegraphics[width=0.95\linewidth]{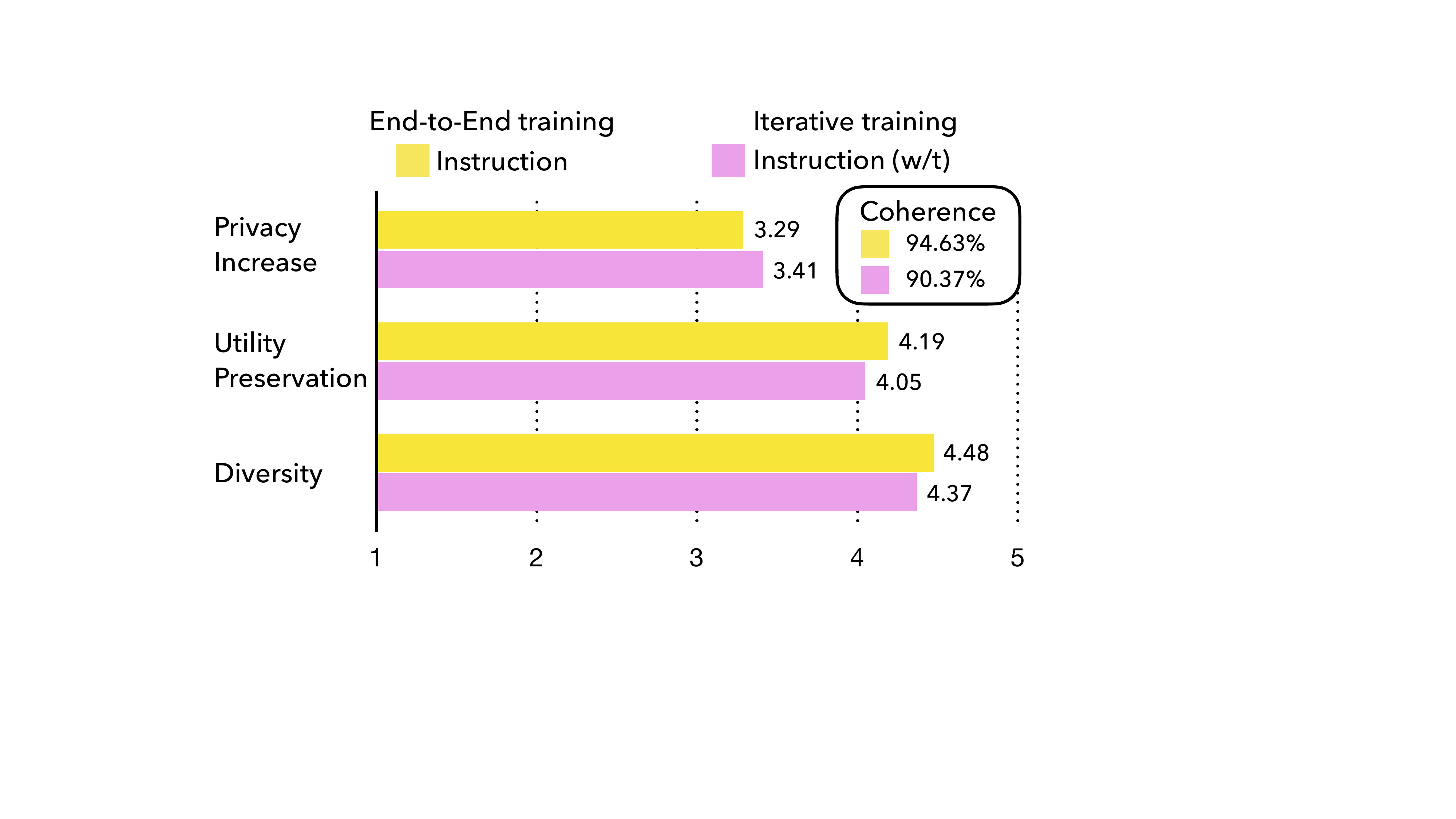}
    \vspace{-5pt}
 \caption{Human evaluation with Likert-scale (1-5) of the top two models. The best model shows moderate privacy increase, high utility preservation, and very high diversity in abstractions. w/t denotes \textit{with thought}.}
    \label{fig:human_eval_likert}
    \vspace{-10pt}
\end{figure}

\vspace{2pt}
\noindent \textbf{Human Evaluation.} We further conduct a human evaluation on 60 sampled self-disclosure test instances. We first use Rank \& Rate \cite{maddela-etal-2023-lens} to rate the abstractions generated from each of the nine models in Table \ref{table:generalization_three_spans_performance} on a scale of 0-100. Each instance receives three ratings from three in-house annotators.
We also report the z-scores, which normalize the raw scores by the mean and standard deviation for each annotator to reduce individual bias. The rank is based on the average z-score. 
For inter-annotator agreement, we calculate an interval Krippendorff's alpha \cite{krippendorff2004reliability} of 0.37 on z-score, and an ordinal Krippendorff's alpha of 0.41 on rank. These values indicate a fair level of agreement, given these methods performing closely, in line with \citet{maddela-etal-2023-lens}.

We then evaluate the top two models on four aspects: privacy increase, utility preservation, and diversity, all rated on a 1-5 Likert scale, along with a binary assessment of coherence, evaluating whether each abstraction integrates seamlessly into the sentence. Detailed definitions for each aspect and Likert scale are provided in Appendix \ref{app:human_eval_likert}.
For inter-annotator agreement, we calculate an interval Krippendorff's alpha, which are 0.46 for Privacy Increase, 0.28 for Utility Preservation, and 0.60 for Diversity, as well as two agree\% (\% of instances where at least two annotators agreed), which are 61.7\%, 80.0\%, 96.7\% for each category respectively, showing a good level of agreement.

Human evaluation results in Table \ref{table:generalization_three_spans_performance} reveal that end-to-end training using the instruction and iterative generation with thought and instruction achieve the highest z-scores, with 0.21 and 0.17 respectively. The additional aspect-based human evaluation conducted on these two models is shown in Figure \ref{fig:human_eval_likert}. Both models are capable of generating high-quality abstractions. The end-to-end training model performs slightly better in utility preservation, diversity, and coherence, while the iterative generation model is better at increasing privacy.

\section{Importance Rating of Self-Disclosures}
\label{sec:importance}
To highlight and abstract self-disclosure more selectively, as users suggested in \S \ref{sec:user_study}, we consider an additional task that rates the importance of each disclosure within context.

\paragraph{Task Definition.} Given the disclosure span and its surrounding context, the task is to estimate how important this disclosure is for others to understand the user's message and communication goals. We consider three levels: \texttt{low}, \texttt{moderate}, and \texttt{high}, corresponding to disclosures that can be removed, essential but can be abstracted, and have to be kept as it is (see App. \ref{app:annotation-guideline} for details). For disclosures that appear in a post's title or body, we consider both the title and body of the post as context. For disclosures in comments (i.e., replies to the main post), the context extends to the entire comment and its parent comment in the reply chain, if existing.

\begin{figure}[!t]
  \centering
    \includegraphics[width=0.95\linewidth]{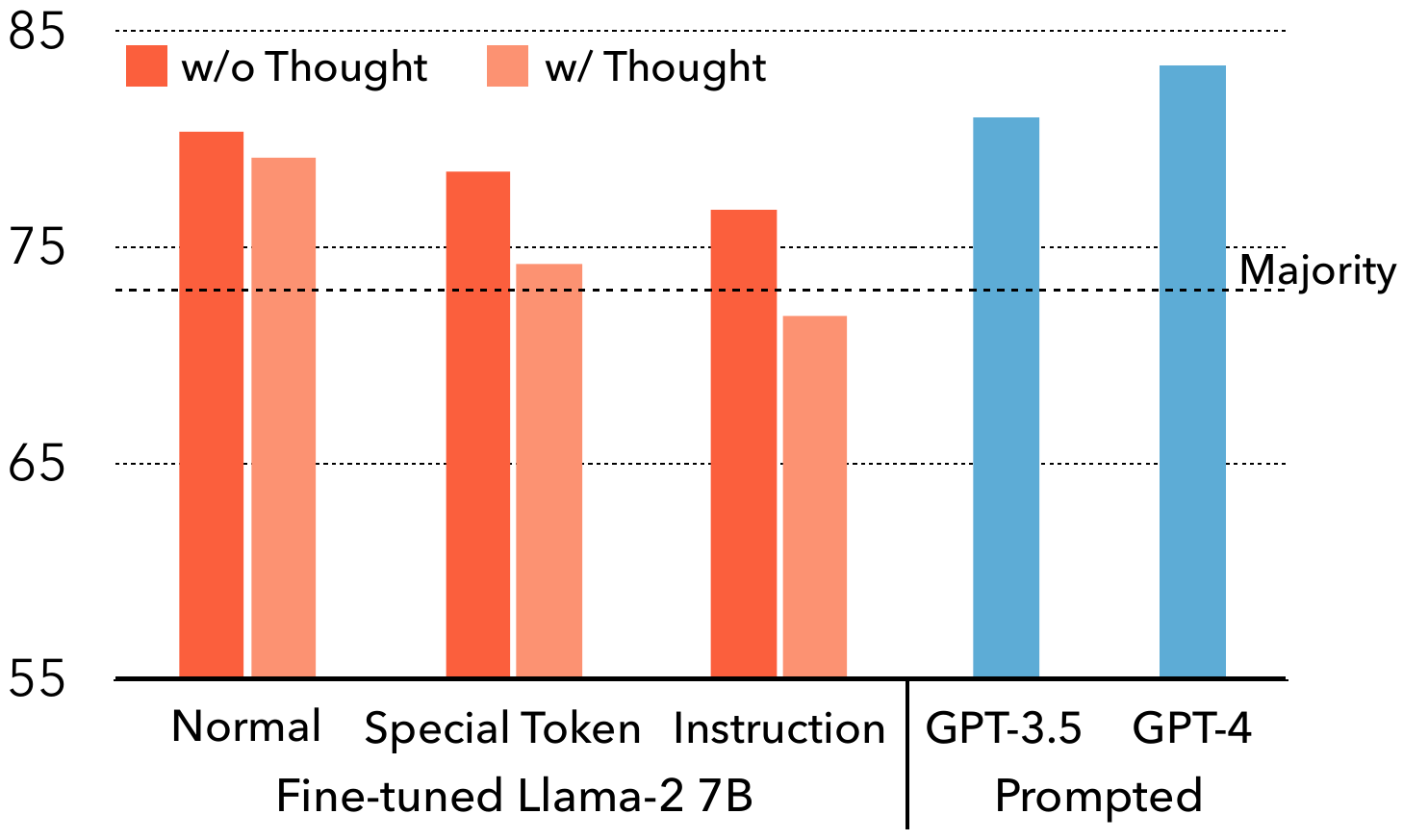}
    \vspace{-5pt}
 \caption{Test results of importance rating, measured in accuracy. Fine-tuning directly on output is better than on thought. The best fine-tuned model achieves comparable performance with GPT-3.5.}
    \label{fig:importance-rating}
    \vspace{-10pt}
\end{figure}

\paragraph{Training Data.} Compared to abstraction, humans perform more effectively than LLMs on importance rating. We use the same train/dev/test split as the abstraction experiment and have each instance annotated by three in-house annotators.
24\% instances reach a consensus and 65\% 
 have agreement between two annotators. Only 11\% exhibit complete disagreement.
 We also calculate Krippendorff's $\alpha$ as 0.29. This fair level of agreement is anticipated given the task's subjective nature. For example, people have different opinions about whether details like age in a dating post should be retained (\textit{``32''}) or abstracted (\textit{``early 30s''}). There often isn't  a clear cut between low and moderate, moderate and high; yet, they provide useful signals to users. Further discussions are in Appendix \ref{app:further_importance_rating_results}. For training labels, we take the majority vote or  \texttt{moderate} if the annotators choose all three levels.

\paragraph{Model.} We fine-tune Llama-2-7B  with various input-output formats as in the abstraction experiment (\S\ref{sec:abstraction}). We use GPT-3.5 to generate the reasoning that leads to the human-assigned ratings.

\paragraph{Evaluation Results.}
Given the task subjectivity, we measure accuracy by considering a prediction as correct if it matches any one of the three annotations. Figure \ref{fig:importance-rating} shows the accuracy of each method in comparison with GPT-3.5 and GPT-4. We find that fine-tuning on thought process degrades performance across all input format performance. The top performing fine-tuned model achieves 80.37\%, on par with GPT-3.5's 80.98\%.

\section{Related Work}
\label{sec:relatedwork}
There is an excellent survey-position paper by \citet{lison-etal-2021-anonymisation}, which has provided a comprehensive review of literature in the NLP for Privacy area. It identified one key research challenge as: ``\textit{Most importantly, they (NLP approaches)
are limited to predefined categories of entities and
ignore how less conspicuous text elements may
also play a role in re-identifying the individual. For
instance, the family status or physical appearance
of a person may lead to re-identification but will
rarely be considered as categories to detect.''} ---  which motivated this very work of ours. We discuss some related works, including the newer ones, below.

\paragraph{Online Self-Disclosure Detection.} Most existing research addressed self-disclosure detection in social media as sentence or document classification, which could not accurately pinpoint the specific disclosure spans. Many prior work focused on one specific kind of self-disclosure, such as medical and mental health conditions \cite{de2016discovering,yates2017depression, benton2017multitask,klein2017detecting, zhao2019identifying,valizadeh2021identifying,valizadeh2023clued}, sexual harassment   \cite{schrading2015analysis,andalibi2016understanding,chowdhury2019speak}, personal opinions or sentiments \cite{cho2022assessing}, and employment history \cite{preoctiuc2015analysis,tonneau2022multilingual}. Other research considered all types of personal information that a user can reveal as a single category \cite{mao2011loose,caliskan2014privacy,bak2014self, balani2015detecting,wang2016modeling,yang2017self,blose2020privacy,reuel2022measuring}. A few works considered self-disclosures of multiple classes, but covered only a small number of categories \cite{lee2023online,akiti2020semantics}, or relied on dictionary/rule-based methods \cite{guarino2022automatic}, or limited to a particular context (e.g., bragging \cite{jin2022automatic} and news comments \cite{umar2019detection}). To address these limitations, we emphasize detecting self-disclosure at the span level and broadening the coverage to include 19 distinct categories. This allows for a more fine-grained detection of privacy leaks for users.

\paragraph{PII Identification and Anonymization.}
Personal Identifiable Information (PII) is closely related to self-disclosures, but with a focus on highly sensitive attributes, such as full names, social security numbers, dates of birth, etc. \cite{regulation2016regulation, morris2022unsupervised, adams2019anonymate, lukas2023analyzing, hathurusinghe2021privacy}. Such sensitive data is more commonly encountered in legal \cite{pilan2022text,mansfield2022behind} and medical text \cite{yue2020phicon, dernoncourt2017identification}, as opposed to in social media or online community sources. Existing tools for PII identification such as Microsoft's Presidio \footnote{\url{https://microsoft.github.io/presidio/}} use regular expressions \cite{subramani2023detecting, mouhammad-etal-2023-crowdsourcing}, and Named Entity Recognition (NER) detectors \cite{honnibal2020spacy}.  However, such approaches indiscriminately mark entities  (e.g., a business phone number on an advertisement) without considering whether the information is self-disclosed. PII anonymization \cite{azurelanguageservice,amazoncomprehend}, widely used in healthcare records management and machine learning training pipelines, replaces sensitive data with masked tokens (e.g., \textit{[xxx]}) or weaker labels (e.g., \textit{[Location]}). This aggressive approach hurts the utility of the message and is not suited for online self-disclosures, which are often voluntary and serve specific functions. To address this, we introduce a novel task of \textbf{self-disclosure abstraction} that strikes a balance of utility and privacy.

\paragraph{Privacy Leakage in Language Models.}

Recent work has shown however that LLMs are prone to leaking personal information \cite{sun2023does,kim2023propile, huang2022large, lukas2023analyzing} and lack the ability to reason about privacy compared with humans \cite{mireshghallah2023can}. This phenomenon of user privacy violation is due to the issue of memorization \cite{carlini2022quantifying, carlini2021extracting}, where LMs recall individual sequences from their pre-training corpora. Recent efforts in solving this leakage problem include differentially-private training \cite{ponomareva2023dp, li2021large}, decoding methods that prevent generation of memorized sequences \cite{ippolito2022preventing}, and prompt self-moderation \cite{chen2023can}. We refer interested readers to the recent surveys of \citet{smith2023identifying}, \citet{ishihara2023training}, and \citet{klymenko2022differential} for additional information on privacy leakage and memorization. Our work takes a different angle by providing a user-centered approach that tackles a root cause of privacy leakage; helping users make more informed decisions when posting online through self-disclosure identification and abstraction. This in turn can help decrease the chance of personal information ending up in pre-training corpora and reduce potential privacy violations.

\section{Conclusion}
We push the first steps in protecting user-side privacy in online self-disclosures.
Our disclosure detection model trained on our new fine-grained corpus with span-level annotations achieves over 65\% of partial span F$_1$ and is further validated through an HCI user study, highlighting its real-world applicability. Responding to the need from participants for balancing privacy risk reduction with message utility, we propose a novel task of self-disclosure abstraction, and explore various fine-tuning methods to generate three diverse abstractions. Our human evaluation shows that the best model can provide diverse abstractions that reduce privacy risks while highly preserving utility. We further fine-tune a model to rate the importance of the self-disclosure on understanding user's perspective and context. This model reaches 80\% accuracy, matching the performance of GPT-3.5, thereby helping users to decide which disclosures to abstract. Overall, we believe our work paves the way for a new direction of using LLMs to protect user-side privacy.

\section*{Limitations}

Our user study (\S \ref{subsec:id_limitations}) has revealed some additional limitations and research directions, including personalization, explainability, and contextual awareness. More specifically, 5 participants suggested that the tool should consider subreddit-specific norms. For example, the \texttt{r/diabetes} subreddit inherently expects that users may discuss their medical condition, rendering some model-predicted highlights redundant. 
Participants also expressed a desire for more transparency in the model’s decision-making process, such as more explanations as to \textit{why} certain highlights were marked as disclosures. Future work could expand our research to include other social platforms to provide broader insights and applicability in diverse social environments.
In this work, we evaluate self-disclosure detection and abstraction individually for accurate assessment. As these tasks work consecutively as a pipeline in practice, future work could conduct user studies on the whole pipeline.
Future research could also investigate the model's performance after quantization, which will allow deploying Llama-7B completely on personal devices for better privacy protection.

\section*{Ethics Statement}
This research was approved by the Institutional Review Board (IRB) at Georgia Institute of Technology.
We take the following measures to safeguard the personal information in our corpus before the annotation process. First, all personal identification information (PII), such as names and emails, is replaced with synthetic data. Second, we hired in-house student annotators (\$18 per hour) instead of crowd workers for annotation. Every annotator was informed that their annotations were being used to create a dataset for online self-disclosure detection. All examples, except for those generated by the model, shown in the paper are synthetic but accurately reflect the real data.
Our user study was approved by IRB at Carnegie Mellon University. The primary data collected from user interviews (including the Reddit posts run through the model) was self-disclosed and gathered in a survey with the participants' consent. In accordance with IRB policy, we anonymized all data collected during the study by removing any PII.
The primary purpose of our models is to provide users with a tool to mitigate the privacy risks associated with online self-disclosures. In cases where the models fail, they do not pose additional risks but rather tend towards overprotection, either by identifying more spans of text or by overly abstracting the disclosed information.
We identified no potential harms that would disproportionately impact marginalized or otherwise vulnerable populations.
To prevent misuse, we will not release our dataset and models to the public. Instead, we will share them upon request with researchers who agree to adhere to these ethical guidelines:
\begin{enumerate}[itemsep=0.2pt, leftmargin=*, topsep=0.2pt, partopsep=0.2pt]
    \item Use of the corpus is limited to research purposes only.
    \item Redistribution of the corpus without the authors' permission is prohibited.
    \item Compliance with Reddit's policy is mandatory.
\end{enumerate}
Annotations of posts that have been deleted by users will be excluded at the time of the data request. To request access, please email the authors.

\section*{Acknowledgments}
The authors would like to thank Srushti Nandu, Chase Perry, and Shuheng Liu for conducting pilot studies; Piranava Abeyakaran, Nour Allah El Senary, Vishnesh Jayanthi Ramanathan, Ian Ligon, Govind Ramesh, Ayush Panda, and Grace Kim for their help with data annotation and evaluation; Yang Chen, Duong Minh Le, and Nghia T. Le for their helpful feedback. We would also like to thank Azure’s Accelerate Foundation Models Research Program graciously providing access to API-based models, such as GPT-4. This research is supported in part by NSF awards IIS-2144493, IIS-2052498 and IIS-2112633, ODNI and IARPA via the ODNI and IARPA via the HIATUS program (contract 2022-22072200004). The views and conclusions contained herein are those of the authors and should not be interpreted as necessarily representing the official policies, either expressed or implied, of ODNI, IARPA, or the U.S. Government. The U.S. Government is authorized to reproduce and distribute reprints for governmental purposes notwithstanding any copyright annotation therein.

\bibliography{anthology,custom}
\bibliographystyle{acl_natbib}

\appendix
\newpage

\section{Self-Disclosure Corpus}
\label{app:corpus}
\subsection{Quality Control}

Recent works have shown that crowdsourcing often leads to lower quality annotations \cite{clark-etal-2021-thats,gilardi2023chatgpt}. To ensure the quality, we hire seven in-house annotators, who are undergraduate students at a US university and native English speakers, and compensate them at a rate of \$18 per hour.  Each annotator first undergoes a training that includes tutorials and 20 exercise examples. During the annotation, we release the data in ten batches, allowing us to constantly monitor performance and provide feedback as needed. We use the BRAT interface for the annotation process.\footnote{\url{https://brat.nlplab.org/}}

\begin{table}[ht!]
  \centering
  \setlength{\tabcolsep}{4pt}
  \resizebox{0.96\columnwidth}{!}{
    \begin{tabular}{lcc}
      \toprule
      \textbf{Category} & \textbf{Krippendorff's $\alpha$} & \textbf{Two Agree (\%)}  \\
      \midrule
      \textsc{Age}  & 0.80  & 73.1 \\
      \textsc{Age\&Gender}  & 0.88 & 81.6 \\
      \textsc{Race/Nationality} & 0.88 & 82.2 \\
      \textsc{Gender} & 0.79 & 74.0 \\
      \textsc{Location} & 0.71 & 65.4 \\
      \textsc{Appearance} & 0.68 & 57.1\\
      \textsc{Wife/GF} & 0.70 & 62.1 \\
      \textsc{Finance} & 0.65 & 57.6 \\
      \textsc{Occupation} & 0.66 & 58.4  \\
      \textsc{Family}  & 0.70 & 61.5  \\
      \textsc{Health} & 0.53 & 46.7 \\
      \textsc{Mental Health} & 0.45 & 37.0 \\
      \textsc{Husband/BF} & 0.78 & 70.8 \\
      \textsc{Education}  & 0.72 & 64.9 \\
      \textsc{Pet} & 0.48 & 37.7  \\
      \textsc{Relationship Status} &  0.48 & 37.3 \\
      \textsc{Sexual Orientation} & 0.58 & 49.9 \\
      \midrule
      \textsc{Overall} &  0.54 & 45.7\\
      \bottomrule
    \end{tabular}}
  \caption{Inter-annotator agreement for each self-disclosure category.}
  \vspace{-12pt}
  \label{table:iaa_by_class}
\end{table}

\subsection{Inter-annotator Agreement}
\label{app:iaa}
Table \ref{table:iaa_by_class} shows the inter-annotator agreement per self-disclosure category, measured by Krippendorff's $\alpha$ \cite{krippendorff2018content} and \textit{Two Agree}. Due to computational
constraints, we calculate Krippendorff's $\alpha$ per post and report the average across the dataset. Two agree is the percentage of words labeled as \textit{disclosure} by both annotators $a1$ and $a2$: $\frac{|\text{Words}_{a1} \cap \text{Words}_{a2}|}{|\text{Words}_{a1} \cup \text{Words}_{a2}|}$. The numbers are higher than or similar to those in other span-level annotation work \cite{dou-etal-2022-gpt,heineman-etal-2023-dancing}.

\section{User Study Recruitment}
\label{app:recruitment}
We first launched a pre-study survey on Prolific, targeting participants who met the following criteria: had an active Reddit account, made at least three posts on Reddit, resided in the U.S., and were 18 years of age or older. The survey asked them to provide posts they had authored and had privacy concerns over. A total of 158 individuals participated in the survey. We kept the participants whose posts are majority text-based. We then invited 21 Reddit users to participate in the interview based on their availability. We make sure of the diverse demographics among the 21 participants where 12 identified as female, 16 were below the age of 50, and 15 held a bachelor's degree or higher.

\section{Further Detection Results}
\label{app:further_results}
\begin{table}[t!]
  \centering
  \setlength{\tabcolsep}{3pt}
  \resizebox{\columnwidth}{!}{
    \begin{tabular}{lcccccc}
      \toprule
      \multirow{ 2}{*}{\textbf{Input}} & \multicolumn{2}{c}{\textbf{Span F$_1$}} & \multicolumn{2}{c}{\textbf{Partial F$_1$}} & \multicolumn{2}{c}{\textbf{Token F$_1$}} \\
      & \textit{Multi} & \textit{Binary} & \textit{Multi} & \textit{Binary} & \textit{Multi} & \textit{Binary} \\
      \midrule
      Normal  &  42.41 & 43.93 & 58.99 & 60.50 & 66.73 & 72.23 \\
      256   & 43.17  & 45.01 & 60.49 & 60.64 & 67.24 & 71.35 \\
      128 & 42.86 & 45.25 & 60.39 & 60.56 &  69.22 & 71.78 \\
      64 & 43.22 & 43.91  & 59.51 & 62.27 & 68.18 & 73.95  \\
      Sentence   & \textbf{48.88} & \textbf{52.92} & \textbf{65.71} & \textbf{72.29} & \textbf{74.17} & \textbf{83.26} \\
      \bottomrule
    \end{tabular}}
  \caption{Test performance of DeBERTaV3-large fine-tuned on various data setups, with training on sentence level achieving the best results. For \textit{Multi}-class models, the results are averaged over all classes, excluding label ``O'', while \textit{Binary} models report F$_1$ for ``disclosure''.}
  \label{table:macro_performance}
\end{table}

In this section, we describe the details of our self-disclosure detection model and show the performance of the binary-class model.

\subsection{Experiment Details}
We fine-tune RoBERTa-large \cite{liu2019roberta},  a
transformer-based encoder with 355M parameters, and DeBERTaV3-large \cite{he2021debertav3} with 435M parameters on our dataset by minimizing the cross-entropy loss for each token’s label. As some words are tokenized into multiple subword tokens, during
inference, we use the hidden states of the first token
to get the label \cite{rei-etal-2022-cometkiwi}.

We experiment with various data processing methods during fine-tuning, given that Reddit posts and comments can significantly vary in length and disclosure spans do not always require extended contexts. Specifically, we segment comments or posts into shorter chunks of 64, 128, and 256 words, as well as individual sentences using Ersatz \cite{wicks-post-2021-unified}.

For self-disclosure classes which are infrequent in Reddit data (specifically \texttt{name} and \texttt{contact} disclosures) and thus insufficient to train models from scratch, we turn to existing models and tools. We use the state-of-the-art NER model, LUKE \cite{yamada-etal-2020-luke} to identify person names and Microsoft Presidio\footnote{\url{https://microsoft.github.io/presidio/}} to recognize contact information such as phone numbers and social media usernames.
To specifically identify self-disclosures as opposed to generic names (e.g., Taylor Swift), we further train a sentence classifier using RoBERTa-large, which achieves 84.4 test F$_1$, to first determine whether a sentence contains a self-disclosure.

\begin{table*}[t!]
  \centering
  \setlength{\tabcolsep}{3pt}
  \resizebox{0.99\textwidth}{!}{
    \begin{tabular}{lccccccccc}
      \toprule
      \multirow{2}{*}{\textbf{Class (\#Spans)}} & \multicolumn{3}{c}{\textbf{RoBERTa-large}} & \multicolumn{3}{c}{\textbf{DeBERTaV3-large}} & \multicolumn{3}{c}{\textbf{GPT-4}} \\
      \cmidrule(lr){2-4} \cmidrule(lr){5-7} \cmidrule(lr){8-10}
      & Span F$_1$ & Partial F$_1$ & Token F$_1$ & Span F$_1$ & Partial F$_1$ & Token F$_1$ & Span F$_1$ & Partial F$_1$ & Token F$_1$ \\
      \midrule
      \textsc{Age (35)}  & 60.87 & 72.46  & \underline{87.04} & 64.62 & 70.77 & 84.97  & \textit{65.71} & \textbf{80.0} & 80.54 \\
      \textsc{Age\&Gender (17)}  & \textit{73.68} &  \textbf{84.21} & \underline{81.63} & 54.05 & 70.27 & 75.47 & 69.77 & 74.42 & 70.97 \\
      \textsc{Race/Nationality (8)} & 60.0 & 60.0 & 64.94 & \textit{82.35} & \textbf{82.35} & \underline{81.25}  & 70.59 & 70.59 & 71.64 \\
      \textsc{Gender (17)} & 61.11 & 61.11  & 56.47 & \textit{72.73} & \textbf{72.73} & 61.73 & 57.14 & 57.14 & \underline{62.50} \\
      \textsc{Location (41)} & 52.87 & 71.26 & 76.60 & \textit{57.78} & \textbf{73.33} & \underline{83.61}
 & 41.30 & 54.35 & 69.63 \\
      \textsc{Appearance (31)} & \textit{44.07} & 64.41 & \underline{78.14} & 35.48 & \textbf{67.74} & 76.27 & 12.77 & 42.55 & 54.71 \\
      \textsc{Wife/GF (30)} & 53.33 & 66.67 & 74.19 & \textit{65.52} & \textbf{75.86} & \underline{78.36} & 48.39 & 64.52 & 74.32 \\
      \textsc{Finance (33)} & 35.82 & 68.66 & 76.57 & \textit{40.0} & \textbf{71.43} & \underline{77.68} & 31.17 & 54.55 & 69.49 \\
       \textsc{Occupation (44)} & 40.0  & 64.44 & 72.39 & \textit{45.65} & \textbf{65.22} & \underline{75.08} & 39.56 & 52.75 & 68.21 \\
       \textsc{Family (44)}  & \textit{50.0} & \textbf{58.70}  & 66.52 & 35.29 & 49.02 & \underline{69.23} & 46.60 & 58.25 & 64.68  \\
      \textsc{Health (40)} & 44.21 & 56.84 & \underline{76.54}   & \textit{45.10} & \textbf{58.82} & 75.94 & 26.45 & 38.02 & 50.17 \\
      \textsc{Mental Health (46)} & 39.22  & \textbf{64.71} & \underline{81.18} & \textit{40.0} & 63.16 & 75.75 & 36.36 & 52.73 & 66.51 \\
      \textsc{Husband/BF (14)} & \textit{68.75} & \textbf{75.0} & \underline{71.91} & 64.71 & 70.59 & 71.11 & 55.17 & 68.97 & 68.49 \\
        \textsc{Education (21)}  & 55.32 &  68.09 & 85.23 & \textit{65.38} & \textbf{69.23} & \underline{86.73}  & 42.55 & 51.06 & 67.37 \\
      \textsc{Pet (15)} & 46.15 & 46.15 & 56.88 & \textit{48.28} & \textbf{55.17} & \underline{61.54} & \textit{48.28} & 48.28 & 56.25 \\
      \textsc{Relationship Status (31)} & \textit{38.36} & 41.10 & 58.80 & 36.92 & \textbf{43.08} & \underline{60.11} & 34.29 & 42.86 & 57.58 \\
      \textsc{Sexual Orientation (12)} & 28.57 & \textbf{76.19} & 68.0 & 25.0  & 58.33 &  66.0 & \textit{52.17} &  69.57 & \underline{71.29} \\
      \midrule
      \textsc{Average} & 48.32 &  64.71 & 72.53 & \textit{48.88}& \textbf{65.71} & \underline{74.17} & 41.93 & 57.68 & 66.14 \\
      \bottomrule
    \end{tabular}}
  \caption{Test performance per class for fine-tuned models and prompted \texttt{GPT-4-0125-preview}. \textit{Italic} highlights the best model for each class on span F$_1$. \textbf{Bold} on partial span $F_1$. \underline{Underline} on token F$_1$.}
  \vspace{-10pt}
  \label{table:performance_by_model_full}
\end{table*}

\subsection{Results}
Table \ref{table:macro_performance} presents the average test-set performance for both binary and multi-class models when fine-tuned under different data configurations. Due to the inherent simplicity, binary models typically outperform their multi-class counterparts.
In addition, we find that dividing the long Reddit posts and comments into shorter pieces generally improves performance. The most significant gain is achieved by segmenting the data at the sentence level, leading to an increase of over 6 span-level F1 points in both binary and multi-class settings, compared to the normal baseline.

Table \ref{table:performance_by_model_full}, a detailed version of Table \ref{table:performance_by_model} from Section \ref{sec:id_model}, shows span, partial span, and token-level F$_1$ of fine-tuned RoBERTa-large, DeBERTaV3-large, and prompted GPT-4 Turbo. For GPT-4 Turbo, we iteratively refined the prompt, listed in Appendix \ref{app:prompt-template}, incorporating definitions, guidelines, and chain-of-thought. We find that generating thought before outputing results leads to an increase of 2.43 partial span F$_1$.
We also discover that when breaking down categories into multiple groups performs worse than detecting all categories at once, as GPT-4 Turbo tend to predict unspecified categories or over-classifiy like ``40 yo man'' as \texttt{appearance}.
For our user study, we use RoBERTa-large model.

\begin{table}[t!]
  \centering
  \setlength{\tabcolsep}{3pt}
  \resizebox{0.99\columnwidth}{!}{
    \begin{tabular}{llll}
      \toprule
      \textbf{Class (\#spans)} & \textbf{Span F$_1$} & \textbf{Partial F$_1$} & \textbf{Token F$_1$} \\
      \midrule
        \textsc{Occupation (75)} & 38.89 (\textcolor{red}{-6.8}) & 61.11 (\textcolor{red}{-4.1}) & 64.14 (\textcolor{red}{-10.9}) \\
      \textsc{Location (31)} & 56.0 (\textcolor{red}{-1.8}) & 66.67 (\textcolor{red}{-6.7})  &  77.35 (\textcolor{red}{-6.3}) \\
            \textsc{Education (10)}  & 57.14 (\textcolor{red}{-8.2})  & 66.67 (\textcolor{red}{-2.6}) & 91.80 (\textcolor{green3}{+5.1})  \\
        \textsc{Relation. Status (6)} & 42.11 (\textcolor{green3}{+5.2}) & 42.11 (\textcolor{red}{-1.0}) & 63.16 (\textcolor{green3}{+3.1}) \\
       \textsc{Family (6)}  & 66.67 (\textcolor{green3}{+31.4}) & 66.67 (\textcolor{green3}{+17.7}) & 74.47 (\textcolor{green3}{+5.2}) \\

      \midrule
      \textsc{Average} & 52.16 (\textcolor{green3}{+4.0}) & 60.64 (\textcolor{green3}{+0.7}) & 74.18 (\textcolor{red}{-0.8})  \\
      \bottomrule
    \end{tabular}}
  \caption{Per-class performance of the fine-tuned DeBERTa in detecting self-disclosure within ShareGPT conversations. Differences compared to in-domain performance are shown in parentheses ().}
  \vspace{-10pt}
  \label{table:performance_sharegpt}
\end{table}

\subsection{ShareGPT}
\label{app:sharegpt}

We describe the process of collecting data and annotating self-disclosures within ShareGPT conversations \footnote{\url{https://huggingface.co/datasets/anon8231489123/ShareGPT_Vicuna_unfiltered}}.
This process is conducted in four steps: \textbf{1)} we filter out conversations turns that are AI-generated, have over 500 tokens, or don't contain ``I'' or ``my'', resulting in 59,533 human-authored turns, \textbf{2)} we randomly select 1,600 conversation turns from them, \textbf{3)} five in-house annotators then identify 105 turns containing self-disclosures, \textbf{4)} these turns undergo a two-round annotation--initial annotation followed by adjudication. In the end, there are 105 human-written turns with annotated self-disclosure spans.

Table \ref{table:performance_sharegpt} presents the performance of DeBERTa, fine-tuned on Reddit data, for each class in detecting self-disclosure within ShareGPT conversations. The model demonstrates good generalizability, performing comparably to its in-domain results.

\section{Further Abstraction Results}
\label{app:further_abstraction_results}
\subsection{Generate a Single Abstraction}
Besides generating three diverse abstractions given a self-disclosure span, we also fine-tune Lllama-2-7B \cite{touvron2023llama} with LoRA \cite{hu2021lora} for generating a single abstraction.
We consider three different \textbf{input formats}, that use standard input, special token, and natural language instructions:

\vspace{3pt}
{
\small
\begin{verbbox}
Sentence:{s}\nDisclosure Span:{d}\nAbstraction
Span:
\end{verbbox}
}
\inputoutput{Standard/Normal Input}{\theverbbox}

\vspace{2pt}

{
\small
\begin{verbbox}
<SENTENCE>{s}<SPAN>{d}<ABSTRACTION>
\end{verbbox}
}
\inputoutput{Special Token}{\theverbbox}

\vspace{2pt}

{
\small
\begin{verbbox}
Your task is to abstract the given disclosure...
Sentence:{s}\nDisclosure Span:{d}
\end{verbbox}
}
\inputoutput{Instruction}{\theverbbox}

For \textbf{output formats}, which is the text that the model is trained to generate and where the loss is calculated, we explore two options. One is solely the desired output, which is the abstraction, and another is a rationale plus the abstraction, also known as chain-of-thought \cite{wei2022chain} training.

\begin{table}[t!]
  \centering
  \setlength{\tabcolsep}{4pt}
  \renewcommand{\arraystretch}{1.2}
  \resizebox{0.9\columnwidth}{!}{
    \begin{tabular}{llcc}
      \toprule
      \multicolumn{2}{c}{\multirow{2}{*}{\textit{BLEU / ROUGE-2}}} & \multicolumn{2}{c}{\textbf{Output}}
    \\
      \cline{3-4}
      \addlinespace[2pt]
      & &  w/o Thought & w/ Thought \\

    \midrule
    \addlinespace[2pt]
      \multirow{3}{*}{\rotatebox[origin=c]{90}{\textbf{Input}}} & Standard & \newcellcolor{red5}{white} 15.3 / 25.1 & \newcellcolor{red6}{black} 14.4 / 22.9 \\
      & Special Token & \newcellcolor{red3}{white} 17.0 / 25.4 &  \newcellcolor{red7}{black} 12.9 / 21.7 \\
      & Instruction & \newcellcolor{red3}{white} 17.9 / 24.8 & \newcellcolor{red2}{white} 18.3 / 25.5  \\
      \bottomrule
    \end{tabular}}
  \caption{Test results on self-disclosure abstraction task. Training with special token and instruction with thought lead to the best performance.}
  \vspace{-10pt}
  \label{table:generalization_one_span_performance}
\end{table}

Table \ref{table:generalization_one_span_performance} reports BLEU \cite{papineni2002bleu} and ROUGE-2 \cite{lin-2004-rouge} for comparing each input and output configurations. We find that using special tokens and instruction with thoughts helps improve performance over the standard method.

\begin{table*}[t!]
  \centering
  \setlength{\tabcolsep}{3pt}
  \resizebox{0.99\textwidth}{!}{
    \begin{tabular}{l}
      \toprule
      \textit{Example 1} \\
      \textbf{Sentence:} {[22M] [21F]} \textbf{My girlfriend} cheated on me with a coworker, it's a little messy. \\
      \textbf{Abstractions:} ``My partner'', ``The individual involved'', ``The person I was in a relationship with'' \\
      \midrule
    
        \textit{Example 2} \\
      \textbf{Sentence:} Did what happened at work make \textbf{my PTSD} from the military worse, or is it a new PTSD? \\
      \textbf{Abstractions:} ``my previous trauma'', ``my mental health condition'', ``my past emotional challenges'' \\
      \midrule
        
        \textit{Example 3} \\
      \textbf{Sentence:} my friend then asked how I could even consider liking eminem cuz of his lyrics, and that \textbf{I am gay}. \\
      \textbf{Abstractions:} ``I identify as LGBTQIA+.'', ``I am part of the rainbow community.'', ``I belong to the LGBTQ+ community.'' \\
      \midrule

      \textit{Example 4} \\
      \textbf{Sentence:} ...then stop and \textbf{I took amitriptyline 10mg once and stop,} then some other sleeping pills... \\
      \textbf{Abstractions:} ``I tried a prescribed medication for a short period of time,'', ``I experimented with a medication for a limited  \\
      time,'', ``I explored the use of a prescribed medication for a brief period,'' \\

        \midrule
        \textit{Example 5} \\
      \textbf{Sentence:} Now I stole and lied to \textbf{my dad and sister} the only two people who go the extra way for me. \\
      \textbf{Abstractions:} ``my closest family members'', ``the people who care about me the most'', ``the individuals who are always \\
      there for me'' \\

        \midrule
        \textit{Example 6} \\
      \textbf{Sentence:} I've been on antidepressants for 6 months and \textbf{I want to kill myself everyday.} \\
      \textbf{Abstractions:} ``I'm feeling a profound sense of despair daily.'', ``I'm experiencing intense feelings of hopelessness every \\
       day.'', ``I'm struggling with extreme sadness and discomfort each day.'' \\

        \midrule
        \textit{Example 7} \\
      \textbf{Sentence:} \textbf{I am an International Student} and i am new to this place. \\
      \textbf{Abstractions:} ``I am a student from another country'', ``I am a foreign student adjusting to my new surroundings'', \\
      ``I am a student coming from another part of the world''\\

        \midrule
        \textit{Example 8} \\
      \textbf{Sentence:} \textbf{I was open about my abusive childhood} and still received a rating for PTSD. \\
      \textbf{Abstractions:} ``I shared my difficult childhood experiences'', ``I spoke openly about my traumatic childhood'', ``I talked  \\
      publicly about my challenging upbringing''\\
      \bottomrule
    \end{tabular}}
  \caption{8 randomly sampled examples with abstractions generated by the fine-tuned Llama-2-7B. The self-disclosure span to abstract is marked in \textbf{bold}.}
  \vspace{-10pt}
  \label{table:abstraction_examples}
\end{table*}

\subsection{Examples}
Table \ref{table:abstraction_examples} displays 8 randomly sampled examples with three abstractions generated by the best fine-tuned Llama-2-7B from Section \ref{sec:abstraction}.

\section{Further Importance Rating Discussions}
\label{app:further_importance_rating_results}
To further illustrate the subjectivity, here is an example where each of three annotators assigned different labels:

\textbf{Post:} “At what age in your life did you want to settle down?”

\textbf{Comment:} “When I was 23 after \textit{I finished my master degree}, I married my best friend.”

\textbf{Disclosure in the comment:} “I finished my master degree,”.

In this example, the annotators have different opinions on whether to keep, abstract or remove the self-disclosure. Each choice reflects a valid perspective: keeping it provides a clear milestone that may resonate with readers; abstracting to “a life milestone” omits specific education details; deleting it as the post only asks for age.

\section{Implementation Details}
\label{app:implementation}
\subsection{GPT-3.5 and GPT-4}
We use gpt-3.5-turbo-0613 as GPT-3.5 and gpt-4-0613 as GPT-4.

\subsection{Evaluation Metrics}
For BLEU \cite{papineni2002bleu}, we use SacreBLEU \cite{post-2018-call}. 
For ROUGE 
\cite{lin-2004-rouge}, we use the one from torchmetrics.\footnote{\url{https://torchmetrics.readthedocs.io/en/stable/text/rouge_score.html}}

\subsection{Experiments}
We implemented our models using Huggingface 4.33.2  \cite{wolf2019huggingface} and PyTorch 2.0.1 \cite{paszke2019pytorch}. All results are from single runs.
\paragraph{Self-disclosure Detection.} We train RoBERTa-large \cite{liu2019roberta} and DeBERTaV3-large \cite{he2021debertav3} on 2 A40 GPUs. We first train on data with single annotations for 10 epochs with batch size of 32, which takes around 16 minutes per run. We evaluate on the dev set  every 50 steps, saving the checkpoint with the highest partial span-level F$_1$. We then fine-tune on a batch of double annotated data for another 10 epochs, which takes about 5 minutes per run. We evaluate every 20 steps on the dev set, and save the checkpoint with the highest partial span-level F1 for final evaluation. We perform a learning rate sweep over 1e-5, 2e-5, 3e-5, 5e-5, 8e-5 on our evaluation set, with 5e-5 being the best for RoBERTa and 3e-5 for DeBERTa. We use AdamW \cite{loshchilov2017decoupled} as the optimizer with weight decay of 0.01. Additionally, we use a cosine learning rate schedule with a warmup ratio of 0.06.

\paragraph{Self-disclosure Abstraction.} We train Llama-2 7B \cite{touvron2023llama} with LoRA \cite{hu2021lora} on 8 A40 GPUs for 5 epochs, with a total batch size of 32, which takes at most 2 hours per run. We use a learning rate of 1e-4, which is the standard when fine-tuning LLMs with LoRA. We use AdamW \cite{loshchilov2017decoupled} as the optimizer with a weight decay of 0.01. Additionally, we use a cosine learning rate schedule with a warmup ratio of 0.03.
For LoRA hyperparameters, we set rank=8, alpha=16, dropout=0.05, target modules=Q and V attention matrices. Additionally, we also update the embedding layer during fine-tuning.

\paragraph{Importance Rating.} We use the same setup and hyperparameters as the abstraction experiment. It takes around 25 minutes per training run.

\section{Annotation Guidelines}
\label{app:annotation-guideline}
\subsection{Self-disclosure Annotation}

Annotators were instructed to annotate explicit self-disclosures that concern the user based on our defined list of categories (Table~\ref{tab:disclosure-categories}). Most of the categories under "attributes" are straightforward to annotate such as the user's age, location, gender, etc. We considered instances where Reddit users revealing both their age and gender in one word, such as \textit{``\sethlcolor{redf1}\hl{M24}''}, under a specific category {\sc Age/Gender} that is different from {\sc Age} and {\sc Gender} which are for disclosures of age and gender individually. For tricky categories, most of which are under "experiences", we provided exact definitions to follow which help the annotators make decisions and ensure consistent labeling. Those definitions are as follows: 

\begin{itemize}[itemsep=2pt,topsep=2pt]
    \item \textit{Appearance} self-disclosures are defined as descriptions of bodily features of the user, such as their height, weight, eye or hair color, or any other specific features.
    
    \item \textit{Health} self-disclosures are defined as the disclosure of a specific disease or illness the user has, specific medications they take, or medical tests they perform.

    \item \textit{Mental Health} self-disclosures are defined as situations where users discuss their feelings, state of mind, or suicidal thoughts.

    \item \textit{Finance} self-disclosures are defined as mentions of specific personal financial details such as details about one's salary, recent transactions, affordability of items, choice of bank, and similar specifics.

    \item \textit{Education} self-disclosures are defined as mentions what the user is currently or planning on studying, or degrees they hold.

    \item \textit{Occupation} self-disclosures are defined as mentions of the current or past occupations of the user.

    \item \textit{Family} self-disclosures are defined as any disclosure that fits within our specified attributes and experiences but concern a family member of the user, such as their parents, siblings, or extended family members.
\end{itemize}

\subsection{Human Evaluation for Abstraction}
\label{app:human_eval_likert}
The following are the 1-5 Likert scales for the aspects: privacy increase, utility preservation, and diversity, used in human evaluation for three-span abstraction (\S\ref{sec:abstraction}).

\vspace{5pt}
\textbf{\textcolor{red3}{Privacy Increase}}:
\begin{enumerate}[label={},itemsep=-2pt,topsep=2pt,leftmargin=10pt]
    \item 1 -- No Privacy Increase: The abstractions are the same or paraphrases to the disclosure span.
    \item 2 -- Low Privacy Increase: The abstractions slightly obscure sensitive details but are still quite similar to the original.
    \item 3 -- Moderate Privacy Increase: The abstractions moderately obscure sensitive details.
    \item 4 -- High Privacy Increase: The abstractions significantly obscure sensitive details and remove details.
    \item 5 -- Maximum Privacy Increase: The abstractions eliminate nearly all specific details.
\end{enumerate}

\vspace{5pt}
\textbf{\textcolor{green3}{Utility preservation}}:
\begin{enumerate}[label={},itemsep=-2pt,topsep=2pt,leftmargin=10pt]
    \item 1 -- No Utility Preserved: The abstractions remove or significantly change the disclosure span, losing all the utility.
    \item 2 -- Low Utility Preserved: The abstractions preserve a small amount of the disclosure span, but major aspects are lost or altered.
    \item 3 -- Moderate Utility Preserved: The abstractions maintain a part of the disclosure span’s utility.
    \item 4 -- High Utility Preserved: The abstractions maintain most of the disclosure span’s utility, with only minor aspects lost.
    \item 5 -- Full Utility Preserved: The abstractions maintain the complete utility of the disclosure span, effectively conveying the intended function.
\end{enumerate}

\vspace{5pt}
\textbf{\textcolor{blue5}{Diversity}}:
\begin{enumerate}[label={},itemsep=-2pt,topsep=2pt,leftmargin=10pt]
    \item 1 - Identical Abstractions: All three abstractions are essentially the same, exhibiting no diversity in wording or style.
    \item 2 - Minimal Diversity: Two of the three abstractions are identical, with only one offering a different expression.
    \item 3 - Low Diversity: All three abstractions are different, yet they exhibit similar styles and only minor variations in wording.
    \item 4 - Moderate Diversity: Each abstraction differs significantly in wording, with about half of the words unique to each. The styles are somewhat varied but maintain a degree of similarity.
    \item 5 - High Diversity: Each abstraction is distinctly unique, both in wording and in expression style, demonstrating a broad range of diversity.
\end{enumerate}

\subsection{Importance Rating Annotation}
\begin{enumerate}[label={},itemsep=-7pt,topsep=0pt,leftmargin=2pt]
    \item \textit{Low} -- Can be removed without compromising the understanding of the user's perspective and context.
    \item \textit{Moderate} -- Adds a meaningful layer to the context but is not essential to the user's overall message. Can be abstracted.
    \item \textit{High} -- Essential for an accurate understanding of the user's perspective and context, should be kept as is.
\end{enumerate}

\section{Prompt Templates}
\label{app:prompt-template}
We first list the prompts used to prompt GPT-3.5 and GPT-4, and then the prompts for fine-tuning the Llama-2-7B model. 
For prompts used in the importance rating experiment (\S\ref{sec:importance}), we provide ones for disclosures in comments; for disclosures in other locations (title, post, subcomments), only the instance part of the prompt is modified.
Please note that within these prompts, we use ``generalization'', ``generalize'', and ``rationale'' as equivalents to ``abstraction'', ``abstract'', and ``thought'' respectively. 

\lstset{
    basicstyle=\ttfamily\footnotesize,
    breaklines=true,
    breakatwhitespace=false,
}

\subsection{GPT-3.5 and GPT-4 Prompts}
\noindent \textbf{Self-disclosure Detection.} We use the following prompt to ask GPT-4 to detect self-disclosures.
\begin{lstlisting}
## Task
Analyze the provided sentence to identify segments containing self-disclosure. Self-disclosure refers to personal information about the author or their close relations.

## Categories
There are 17 specific categories to consider:
* Age: "I am a 23-year-old"
* Gender: "I'm just a girl"
* Age_Gender: only when age and gender are combined in a *single* word, such as "20F" or "32M".
* Sexual_Orientation: "I'm a straight man"
* Race_Nationality: mentions of the user's nationality/race/ethnicity.
* Wife_GF: disclosures indicating the author has a wife, girlfriend, or fiancee, such as "My gf."
* Husband_BF: disclosures indicating the author has a husband, boyfriend, or fiance, such as "My bf".
* Relationship_Status: only includes mentions of marital status, being in a romantic relationship, or being single. For example, "my partner".
* Family: mentions of specific family members, as well as disclosures that related to themselves, such as "My child is 3 year old".
* Health: includes a wide range of health-related information, from discussing specific diseases or conditions to mentioning medications, medical tests, or treatments.
* Mental_Health: includes a broad range of emotional states and feelings, not necessarily limited to specific mental health diagnoses. Examples include feelings of sadness, anxiety, stress, or comments about struggling with emotional well-being in general.
* Location: captures specific geographical details disclosed by the author. It includes precise locations such as addresses, cities, countries, or distinctive landmarks, so vague spans like "this place" should not be categorized.
* Appearance: "I am 6'2".
* Pet: "I have two musk turtles"
* Occupation: "I'm a motorcycle tourer (by profession)", student should be categorized as Education.
* Education: "I got accepted to UCLA"
* Finance: any financial situations, not necessarily exact amounts.

## Sentence to Analyze
"{sentence}"

Please note:
1. Focus only on disclosures related to the author or their close relations. Avoid marking disclosures about others.
2. For disclosures about close relations, like family members, partners, pets, you should tag them with the subject category (e.g., "Family"). For instance, "My sister is pregnant" should be tagged as "Family" rather than splitting into "Family" and "Health".
3. Always include subjects or pronouns that are adjacent to the disclosure in the span.
4. Never include temporal details in the span.
5. Avoid tagging hypothetical or non-factual information.
6. Do not include non-informational word like "also" in the span.
7. Health and Mental_Health disclosures can be general, do not have to be explicit.

Read the sentence carefully and identify all the segments that fit these categories. Write down a brief thought process first, with the 7 notes in mind. Then finalize the self-disclosure spans in a JSON format: the key must be the self-disclosure span and the value must be its corresponding category.
\end{lstlisting}

\vspace{2pt}
\noindent \textbf{Abstraction Comparison in Section \ref{sec:abstraction}.}
The following zero-shot prompts are used to instruct GPT-4 to compare sentence paraphrasing, sentence, and span-level abstraction.

* Sentence Paraphrasing:
\begin{lstlisting}
Your task is to generate a paraphrase for the given sentence.

Sentence: "{sentence}"
Paraphrased Sentence:
\end{lstlisting}

* Sentence-level Abstraction (without disclosure spans provided):
\begin{lstlisting}
Your task is to generate a paraphrase that conveys similar meaning and maintains the same utlity while removing any overly specific or sensitive information for the given sentence.

Sentence: "{sentence}"
Paraphrased Sentence (Reducing Private Information):
\end{lstlisting}

* Sentence-level Abstraction (with disclosure spans provided):
\begin{lstlisting}
Your task is to generate a paraphrase that conveys similar meaning and maintains the same utlity while removing overly specific or sensitive information for the given sentence. Besides the sentence, I will give you a list of disclosures that need to be addressed.

Sentence: "{sentence}"
Disclosures: {disclosures}
Paraphrased Sentence (Reducing Private Information):
\end{lstlisting}

* Span-level Abstraction:
\begin{lstlisting}
Your task is to generalize the given 'disclosure span' in the sentence. Provide a generalized span that conveys similar meaning but remove any overly specific or sensitive information.

Remember the following criteria:
* Only the disclosure span should be generalized; the rest of the sentence should remain intact.
* Generalized span should retain the essence and utility of the original span.
* Make sure the generalized span fits seamlessly into the original sentence, maintaining proper syntax and grammar.

Sentence: "{sentence}"
Disclosure Span to Revise: "{span}"
Generalized Span:
\end{lstlisting}

\vspace{2pt}
\noindent \textbf{Self-disclosure Abstraction.} We use the following prompt to ask GPT-3.5 (train/dev sets) or GPT-4 (test) to generate three abstraction spans with thought process for each data instance.
\begin{lstlisting}
Your task is to generalize the given 'disclosure span' in the sentence. Provide three diverse generalized spans that convey similar meaning but remove any overly specific or sensitive information.

Example 1:
Sentence: "Should I submit a 1470 SAT score to Carnegie Mellon and Duke?"
Disclosure Span to Revise: "1470 SAT score"
Rationale: The original disclosure span explicitly states a specific SAT score, which may be sensitive information for some individuals. To retain the context that the score is competitively high while removing the exact number, we can use generalized terms.
Generalized Spans: {{"span 1": "a high 1400-range SAT score", "span 2": "an SAT score in the upper 1400s", "span 3": "an SAT score above 1450"}}

Example 2:
Sentence: "My Top Spotify Artists of 2022 were Charli XCX, Taylor Swift, Carly Rae Jepsen, Snow Strippers, and Danny L Harle."
Disclosure Span to Revise: "Top Spotify Artists of 2022 were Charli XCX, Taylor Swift, Carly Rae Jepsen, Snow Strippers, and Danny L Harle."
Rationale: The original span lists very specific artists that the person listened to in 2022, which can be considered personal and revealing. To maintain the general idea that the person has favorite artists from that year without giving away the exact names, generalized terms can be used.
Generalized Spans: {{"span 1": "favorite artists of 2022 included a mix of pop and electronic musicians", "span 2": "top artists from 2022 covered a range of styles", "span 3": "favorite artists last year were from various genres"}}

Example 3:
Sentence: "I just turned 32 last month and have been really reflecting on my relationships with people from my relationship with my siblings to coworkers."
Disclosure Span to Revise: "just turned 32 last month"
Rationale: The original span provides very specific details about the person's age and the timing of their birthday, which could be seen as personal information. To keep the essence of the span-that the individual is in their early 30s and recently had a birthday-while making it less specific, generalized terms can be used.
Generalized Spans: {{"span 1": "recently entered my early 30s", "span 2": "turned into my early thirties not long ago", "span 3": "just started my third decade a little while back"}}

First, provide a rationale explaining why the disclosure span needs generalization. Then, offer three diverse generalized alternatives in a JSON format like this: {{"span 1": "xxx", "span 2": "xxx", "span 3": "xxx"}}.

Criteria:
* Only the disclosure span should be generalized; the rest of the sentence should remain intact.
* Generalized spans should be diverse but should all retain the essence of the original span.
* Make sure the generalized span fits seamlessly into the original sentence, maintaining proper syntax and grammar.

Sentence: "{sentence}"
Disclosure Span to Revise: "{span}"
Rationale:
\end{lstlisting}


\vspace{2pt}
\noindent \textbf{Importance Rating.}
\begin{lstlisting}
Rate the importance of the disclosure span in a Reddit comment within the context into three Likert-scale:
* Low Importance: Can be removed without compromising the understanding of the commenter's perspective and context.
* Moderate Importance: Adds a meaningful layer to the context but is not essential to the commenter's overall message. Can be generalized.
* High Importance: Essential for an accurate understanding of the commenter's perspective and context, should be kept as is.

Given instance:
* Title: {title}
* Post: {post}
* Comment: {comment}
* Disclosure Span in the Comment: {disclosure}
{post_empty_explaination}
Note: The disclosure span is marked between special tokens <disclosure></disclosure> in the Comment.

Read the Title, Post, and Comment carefully to understand the context. Write down your thought process. And in the end, provide your importance rating of the disclosure span in a JSON format: {{"Importance": "Low/Moderate/High"}}.
\end{lstlisting}

The following prompt is used to prompt GPT-3.5 to generate thought process for the human-annotated importance rating:
\begin{lstlisting}
Provide the thought process of the importance rating of a disclosure span in a Reddit comment.

Here are the definitions of the three importance rating scales:
* Low Importance: Can be removed without compromising the understanding of the commenter's perspective and context.
* Moderate Importance: Adds a meaningful layer to the context but is not essential to the commenter's overall message. Can be generalized.
* High Importance: Essential for an accurate understanding of the commenter's perspective and context, should be kept as is.

Given instance:
* Title: {title}
* Post: {post}
* Comment: {comment}
* Disclosure Span in the Comment: {disclosure}
* Human Rating of Importance: {human_rating}
{post_empty_explaination}
Note: The disclosure span is marked between special tokens <disclosure></disclosure> in the Comment.

Instructions:
1. Carefully read the Title, Post, and Comment to understand the context.
2. Based on the human rating, write a detailed thought process that leads to this rating, according to the definitions. Your thought process should build up the reasoning that culminates in the rating, rather than stating the rating and then explaining it.
3. Ensure that your thought process is clear and straight to the point, avoiding filler sentences or unnecessary elaboration.
4. Present your thought in a JSON format of {{"Thought": "xxx"}}.
\end{lstlisting}

\subsection{One-Span Abstraction}
\noindent \textbf{Instruction:}
\begin{lstlisting}
Your task is to generalize the given 'disclosure span' in the sentence, which is providing a generalized alternative that is less specific but retains the core meaning of the original span.

Remember the following criteria:
* Only the disclosure span should be generalized; the rest of the sentence should remain intact.
* Make sure the generalized span fits seamlessly into the original sentence, maintaining proper syntax and grammar.

Sentence: "{sentence}"
Disclosure Span to Revise: "{span}"
Generalized Span:
\end{lstlisting}

\noindent \textbf{Instruction with thought:}
\begin{lstlisting}
Your task is to generalize the given 'disclosure span' in the sentence.

Please follow these steps:
1. First, provide a rationale explaining why the disclosure span needs generalization. 
2. Then, provide a generalized alternative that is less specific but retains the core meaning of the original span.

Remember the following criteria:
* Only the disclosure span should be generalized; the rest of the sentence should remain intact.
* Make sure the generalized span fits seamlessly into the original sentence, maintaining proper syntax and grammar.

Sentence: "{sentence}"
Disclosure Span to Revise: "{span}"
Rationale:
\end{lstlisting}

\subsection{Three-Span Abstraction}

\noindent \textbf{End-to-end instruction:}
\begin{lstlisting}
Your task is to generalize the given 'disclosure span' in the sentence. Provide three diverse generalized spans that convey similar meaning but remove any overly specific or sensitive information.

Remember the following criteria:
* Only the disclosure span should be generalized; the rest of the sentence should remain intact.
* Generalized spans should be diverse but should all retain the essence of the original span.
* Make sure the generalized span fits seamlessly into the original sentence, maintaining proper syntax and grammar.
* Provide three diverse generalized alternatives in a JSON format like this: {{"span 1": "xxx", "span 2": "xxx", "span 3": "xxx"}}.

Sentence: "{sentence}"
Disclosure Span to Revise: "{span}"
Generalized Spans:
\end{lstlisting}

\noindent \textbf{End-to-end instruction with thought:}
\begin{lstlisting}
Your task is to generalize the given 'disclosure span' in the sentence. Provide three diverse generalized spans that convey similar meaning but remove any overly specific or sensitive information.

Please follow these steps:
1. First, provide a rationale explaining why the disclosure span needs generalization. 
2. Then, offer three diverse generalized alternatives in a JSON format like this: {{"span 1": "xxx", "span 2": "xxx", "span 3": "xxx"}}.

Remember the following criteria:
* Only the disclosure span should be generalized; the rest of the sentence should remain intact.
* Generalized spans should be diverse but should all retain the essence of the original span.
* Make sure the generalized span fits seamlessly into the original sentence, maintaining proper syntax and grammar.

Sentence: "{sentence}"
Disclosure Span to Revise: "{span}"
Rationale:
\end{lstlisting}

\noindent \textbf{Iterative instruction:}
\begin{lstlisting}
Your task is to generalize the given 'disclosure span' in the sentence, which is providing a generalized alternative that is less specific but retains the core meaning of the original span.

Remember the following criteria:
* Only the disclosure span should be generalized; the rest of the sentence should remain intact.
* The generalized span should be different from the example generalizations but should retain the essence of the original span.
* Make sure the generalized span fits seamlessly into the original sentence, maintaining proper syntax and grammar.

Sentence: "{sentence}"
Disclosure Span to Revise: "{span}"
Example Generalizations: {examples}
Generalized Span:
\end{lstlisting}

\noindent \textbf{Iterative instruction with thought:}
\begin{lstlisting}
Your task is to generalize the given 'disclosure span' in the sentence.

Please follow these steps:
1. First, provide a rationale explaining why the disclosure span needs generalization. 
2. Then, offer one diverse generalized alternatives that is different from the example generalizations provided.

Remember the following criteria:
* Only the disclosure span should be generalized; the rest of the sentence should remain intact.
* The generalized span should be different from the examples but should retain the essence of the original span.
* Make sure the generalized span fits seamlessly into the original sentence, maintaining proper syntax and grammar.

Sentence: "{sentence}"
Disclosure Span to Revise: "{span}"
Example Generalizations: {examples}
Rationale:
\end{lstlisting}

\subsection{Importance Rating}
\noindent \textbf{Instruction:}
\begin{lstlisting}
Rate the importance of the disclosure span in a Reddit comment within the context into three Likert-scale:
* Low Importance: Can be removed without compromising the understanding of the commenter's perspective and context.
* Moderate Importance: Adds a meaningful layer to the context but is not essential to the commenter's overall message. Can be generalized.
* High Importance: Essential for an accurate understanding of the commenter's perspective and context, should be kept as is.

Given instance:
* Title: {title}
* Post: {post}
* Comment: {comment}
* Disclosure Span in the Comment: {disclosure}
{post_empty_explaination}
Note: The disclosure span is marked between special tokens <disclosure></disclosure> in the Comment.

Read the Title, Post, and Comment carefully to understand the context, and provide your importance rating of the disclosure span.
\end{lstlisting}

\vspace{2pt}
\noindent \textbf{Instruction with thought:}
\begin{lstlisting}
Rate the importance of the disclosure span in a Reddit comment within the context into three Likert-scale:
* Low Importance: Can be removed without compromising the understanding of the commenter's perspective and context.
* Moderate Importance: Adds a meaningful layer to the context but is not essential to the commenter's overall message. Can be generalized.
* High Importance: Essential for an accurate understanding of the commenter's perspective and context, should be kept as is.

Given instance:
* Title: {title}
* Post: {post}
* Comment: {comment}
* Disclosure Span in the Comment: {disclosure}
{post_empty_explaination}
Note: The disclosure span is marked between special tokens <disclosure></disclosure> in the Comment.

Read the Title, Post, and Comment carefully to understand the context. Write down your thought process. And in the end, provide your importance rating of the disclosure span in a JSON format: {{"Importance": "Low/Moderate/High"}}.
\end{lstlisting}

\end{document}